\documentclass[twoside]{article}

\usepackage[accepted]{aistats2025}

\usepackage{microtype}
\usepackage{graphicx}
\usepackage{subfigure}
\usepackage{booktabs} 
\usepackage{adjustbox}
\usepackage{wrapfig}
\usepackage{multirow}
\usepackage[page,header]{appendix}
\usepackage{titletoc}
\usepackage{minitoc}
\usepackage{amssymb}

\usepackage[dvipsnames]{xcolor}
\usepackage{tikz}
\usetikzlibrary{fit, tikzmark, calc}
\newcommand\drawCodeBox[3]{%
    \begin{tikzpicture}[remember picture,overlay]
        \coordinate (start) at ([yshift=1.7ex]pic cs:#2);
        \coordinate (end) at ([yshift=-0.8ex]pic cs:#3);
        \node[inner sep=2pt,#1,fit=(start) (end),rectangle, rounded corners=1mm, fill opacity=0.2] {};
    \end{tikzpicture}%
}

\usepackage{hyperref}
\usepackage{enumitem}
\usepackage[utf8]{inputenc} 
\usepackage[T1]{fontenc}    
\usepackage{hyperref}       
\usepackage{url}            
\usepackage{booktabs}       
\usepackage{amsfonts}       
\usepackage{nicefrac}       
\usepackage{microtype}      
\usepackage{xcolor}         
\usepackage{algorithm}
\usepackage{algorithmic}
\usepackage{amsmath}
\usepackage[capitalize,noabbrev]{cleveref}
\usepackage{color}

\newtheorem{theorem}{Theorem}[section]
\newtheorem{proposition}{Proposition}[section]
\newtheorem{lemma}{Lemma}[section]
\newtheorem{corollary}{Corollary}[section]
\newtheorem{assumption}{Assumption}[section]

\newtheorem{definition}{Definition}[section]

\newtheorem{property}{Property}[section]

\newtheorem{corollary*}{Corollary}
\newtheorem{lemma*}{Lemma}
\newtheorem{theorem*}{Theorem}
\newcommand{\bs}[1]{\boldsymbol{#1}}

\usepackage[round]{natbib}

\begin{document}

%

\runningtitle{On the Power of Adaptive Weighted Aggregation}

%
\runningauthor{Dun Zeng, Zenglin Xu, Shiyu Liu, Yu Pan, Qifan Wang, Xiaoying Tang}

\twocolumn[
\aistatstitle{On the Power of Adaptive Weighted Aggregation \\ in Heterogeneous Federated Learning and Beyond}
\vspace{-3mm}
\aistatsauthor{Dun Zeng \And Zenglin Xu$^*$ \And Shiyu Liu}
\aistatsaddress{University of Electronic Science \\ and Technology of China \\ Chengdu, China \And Fudan University \\ Shanghai, China \And University of Electronic Science \\ and Technology of China \\ Chengdu, China} 
\vspace{-3mm}
\aistatsauthor{Yu Pan \And Qifan Wang \And Xiaoying Tang}
\aistatsaddress{Harbin Institute of Technology \\ Shenzhen, China \And Meta AI \\ Menlo Park, USA \And Chinese University of Hong Kong \\ Shenzhen, China} 
]

\begin{abstract}
Federated averaging (FedAvg) is the most fundamental algorithm in Federated learning (FL). Previous theoretical results assert that FedAvg convergence and generalization degenerate under heterogeneous clients. 
However, recent empirical results show that FedAvg can perform well in many real-world heterogeneous tasks.
These results reveal an inconsistency between FL theory and practice that is not fully explained. 
In this paper, we show that common heterogeneity measures contribute to this inconsistency based on rigorous convergence analysis.
Furthermore, we introduce a new measure \textit{client consensus dynamics} and prove that \textit{FedAvg can effectively handle client heterogeneity when an appropriate aggregation strategy is used}. 
Building on this theoretical insight, we present a simple and effective FedAvg variant termed FedAWARE. 
Extensive experiments on three datasets and two modern neural network architectures demonstrate that FedAWARE ensures faster convergence and better generalization in heterogeneous client settings. Moreover, our results show that FedAWARE can significantly enhance the generalization performance of advanced FL algorithms when used as a plug-in module.
\end{abstract}

\section{Introduction}

Federated Learning (FL) is an emerging distributed training paradigm~\citep{kairouz2021advances}, where a central server orchestrates multiple clients jointly to optimize a machine learning model. The pioneering algorithm, federated averaging (FedAvg)~\citep{mcmahan2017communication}, provides the general local-update framework. It only requires infrequent communication between a server and clients. Thus, FedAvg is especially suitable for FL settings where communication costs are a major bottleneck. FedAvg's simplicity and empirical effectiveness made it the basis of almost all subsequent federated optimization algorithms.

However, the convergence and generalization performance of FedAvg is hindered by client heterogeneity. Specifically, as local updates on clients become more diverse (i.e., as data heterogeneity increases), FedAvg may require more communication rounds to converge. Additionally, this process often results in unstable generalization performance, as seen in the fluctuations of test accuracy (the ``spikes'' problem).
To understand the effects of heterogeneity, plenty of theoretical results~\citep{zhao2018federated, karimireddy2020scaffold, jhunjhunwala2022fedvarp} provided a clear explanation via rigorous convergence analysis. 
These results often use common heterogeneity measures, such as bounded gradient dissimilarity~\citep{li2020federated}, and implicitly assume that federated learning is continuously impacted by worst-case heterogeneity.
However, this theoretical understanding does not always align with empirical findings~\citep{reddi2020adaptive, charles2021large, wu2023anchor, wang2024on}, which show that FedAvg can perform comparably to advanced FL method~\citep{karimireddy2020scaffold} in real-world heterogeneous tasks~\citep{caldas2018leaf}. This discrepancy may arise because FedAvg only encounters severe heterogeneity in a small fraction of the training process~\citep{wang2024on}.
These results highlight an inconsistency between FL theory and practice. Understanding this gap is crucial for designing and evaluating future federated algorithms. This motivates the question \textit{are we overlooking key properties that could explain why FedAvg and its variants perform well in practice under general heterogeneous conditions?}

We intend to answer the question by revisiting the training dynamics of FedAvg via convergence analysis. Firstly,
to accurately describe the heterogeneity impacts, we introduce a concept of \textit{client consensus dynamics}, defined as the cumulative expected norm of aggregated local updates over the communication rounds. Through rigorous theoretical analysis, we show that \textit{the impact of data heterogeneity on FL training can be mitigated if the attained client consensus dynamics is small} (see Theorem~\ref{theorem:fedavg_consensus}).
This suggests that coordinating heterogeneity impacts through carefully designed server-side aggregation could lead to faster convergence of FL training, as shown in Corollary~\ref{cor:adaptive_fedavg}.
Furthermore, we induce a heterogeneity measure, \textit{local update diversity}, to quantify the divergence between local and global updates in practice. We state that \textit{FL is amenable to better generalization if the training procedure attains high local update diversity dynamics}.

Based on these analyses, we propose a simple and effective variant of FedAvg termed \underline{Fed}erated \underline{A}daptive \underline{W}eighted \underline{A}gg\underline{RE}gator (Fed\textbf{AWARE}). Through intensive experiments on training modern neural networks using image and text datasets, we demonstrate that FedAWARE achieves faster convergence and better generalization than most baselines under heterogeneous clients. Furthermore, the AWARE module works with existing federated algorithms, significantly enhancing their generalization performance in data heterogeneity by improving the local update diversity dynamics. This paper contributes key insights into the convergence of FedAvg variants. The results can guide the design and evaluation of future federated algorithms in heterogeneous optimization.


\section{Preliminaries \& Related Works} \label{sec:prel}

We consider a standard cross-device FL task~\citep{kairouz2021advances}, which minimizes a finite sum of local empirical objectives:
$$
\min_{\bs{x} \in \mathcal{X}} f(\bs{x}) := {\sum}_{i=1}^N \bs{\lambda}_i f_i(\bs{x}) := {\sum}_{i=1}^N\bs{\lambda}_i \mathbb{E}_{\xi_i \sim \mathcal{D}_i}[F_i(\bs{x}, \xi_i)],
$$
where $\bs{x} \in \mathcal{X} \subseteq \mathbb{R}^d$ is parameters of machine learning model, $f(\bs{x})$ is the global objective weighted by $\bs{\lambda}_i (s.t., \sum_{i=1}^N\bs{\lambda}_i = 1$, $ \bs{\lambda}_i\geq0, \forall i)$, $\xi_i$ is stochastic batch data, and $\mathcal{D}_i$ denotes dataset on the $i$-th client ($i\in \{1,2,\ldots,N$\}). 
The FedAvg minimizes the global objective involving alternative client optimization and server optimization procedures~\citep{reddi2020adaptive, mcmahan2017communication}:
\begin{equation}\label{eq:framework}
\begin{aligned}
& \text{\textbf{Client: }}\bs{g}_i^{t} = \bs{x}_i^{t, K} - \bs{x}_i^{t, 0} = \eta_l {\sum}_{k=0}^{K-1} \nabla F_i(\bs{x}_i^{t, k});\\
& \text{\textbf{Server: }}\bs{x}^{t+1} = \bs{x}^{t}-\eta_g {\sum}_{i=1}^N \bs{\lambda}_i \bs{g}_i^{t} = \bs{x}^{t}- \eta_g \boldsymbol{G}^t
\end{aligned}
\end{equation}
where $\nabla F_i(\bs{x})$ denotes stochastic gradients over a mini-batch of samples, $\boldsymbol{G}^t$ is the \textit{pseudo-gradient} for global gradient descent, $\bs{x}_i^{t, k}$ denotes client $i$’s model after the $k$ local update steps at the $t$-th communication round, and $\eta_l$ is the client learning rate. 
The federated optimization framework covers a broader range of subsequent FL algorithms~\citep{mcmahan2017communication, reddi2020adaptive, wang2022communication, wang2021field}. 

Our goal is to train a robust and well-generalized global machine-learning model under the potential negative impacts of heterogeneous clients. Therefore, this work is related to previous studies on heterogeneous federated optimization and adaptive aggregation strategies.

\textbf{Heterogeneous federated optimization }
The basic federated optimization algorithm, FedAvg~\citep{mcmahan2017communication} significantly reduces communication costs. Subsequent works built upon FedAvg to address challenges related to convergence guarantees and heterogeneity issues. For example, some approaches introduced a regularization term in the client objectives~\citep{li2020federated}, while others incorporated server momentum~\citep{hsu2019measuring}. Several studies have analyzed the convergence rate of FedAvg and demonstrated its degradation with system heterogeneity~\citep{li2020federated, wang2019adaptive} and statistical heterogeneity~\citep{zhao2018federated, khaled2019first}. \textsc{SCAFFOLD}~\citep{karimireddy2020scaffold} utilizes control variates to mitigate client drift and achieve convergence rates independent of the level of heterogeneity. \textsc{FedNova}~\citep{wang2020tackling} addresses objective inconsistency issues arising from system heterogeneity through local update regularization. Besides, adaptive methods~\citep{zaheer2018adaptive, reddi2019convergence, xie2019local} have proven effective in non-convex optimizations. In the context of federated optimization, FedYogi~\citep{reddi2020adaptive, zaheer2018adaptive} and FedAMS~\citep{acar2020federated} are representative adaptive federated optimization algorithms that incorporate Adam-like momentum and adaptive terms to address heterogeneity issues. For more detailed comparisons, we refer to the survey~\citep{kairouz2021advances}. These related works demonstrate the ongoing efforts to address heterogeneity issues in FL.

\textbf{Adaptive weighting in FL }
The aggregation weights typically represent the importance of each local function in the FL global objective. Local function reweighting scheme has been adopted to improve the fairness~\citep{li2019fair, mohri2019agnostic}, robustness~\citep{li2020tilted}, and generalization~\citep{li2023revisiting} via adjusting $\lambda$ for FedAvg. However, existing adaptive weighting strategies are still based on heuristics. They typically assign a score to each client based on the local dataset properties~\citep{mcmahan2017communication, zhao2024fedsw, ye2023feddisco, li2023revisiting}, local empirical loss~\citep{li2019fair,li2020tilted} or local updates information~\citep{chen2024fedawa}. Then, the aggregation of each client depends on the normalized score among a certain set of clients. Despite these methods being empirically efficient, they provide no explicit theoretical objective for the weight design. In contrast, FedAWARE's non-heuristic weight strategy relies on a clear objective with a closed-form solution. Moreover, we present a unified convergence analysis, and the results cover a broader range of FedAvg variants.

\section{Proposed Algorithm: FedAWARE}\label{sec:fedaware}

We present the details of FedAWARE in Algorithm~\ref{alg:gdm}. It suggests the FL server conducts gradient descent with a simple \textit{adaptive averaging} of \textit{moving-averaged} local updates $\bs{d}^t = \sum_{i=1}^N \bs{\lambda}_i^t \bs{m}_i^t$ by solving
\begin{equation}\label{eq:norm_minimize}
\bs{\lambda}^t = \min_{\bs{\lambda}}\Big\{\big\|\sum_{i=1}^N \bs{\lambda}_i \bs{m}_i^t \big\|^2\big| \sum_{i=1}^N \bs{\lambda}_i = 1, \bs{\lambda}_i \geq 0 \; \forall i\Big\},
\end{equation}
where $\|\cdot\|$ denotes $\ell_2$ norm and $\bs{m}_i^t$ is moving-averaged local updates $\bs{g}_i^t$ for the $i$-th client at communication round $t$. Specifically, this algorithm involves two vital components. FedAWARE can also work with other advanced federated algorithms.

\textbf{Moving-averaged local updates supports for partial client participation  }
We use historical moving-averaged local updates to approximate the local updates of the nonparticipants at a round. In detail, we use a coefficient $\alpha$ to control the approximation with update rules as:
\begin{equation}\label{eq:momentum_rule}
\bs{m}_i^{t} = \begin{cases} (1-\alpha) \bs{m}_i^{t-1} + \alpha \bs{g}_i^t, & \text{if}\; i \in S^t\\ \bs{m}_i^{t-1}, & \text{if} \; i \notin S^t,
\end{cases}
\end{equation}
where $S^t$ is a set of selected clients at round $t$. Moreover, the moving-averaged local 
updates can further stabilize the degree of local heterogeneity as discussed in momentum-based FL methods~\citep{hsu2019measuring}.

\begin{algorithm}[t]
\caption{\textbf{FedAWARE}}
\label{alg:gdm}
\begin{algorithmic}[1]
\REQUIRE $\bs{x}^0, \bs{m}^0, \alpha$
\FOR{round $t$ $\in$ $[T]$}
    \STATE Server sample clients $S^t$ and broadcast $\bs{x}^t$
    \FOR{client $i$ $\in$ $S^t$ in parallel}
        \STATE $\bs{x}^{t,0}_i = \bs{x}^t$
        \FOR{local update step $k$ $\in$ $[K]$}
            \STATE $\bs{x}^{t,k}_i = \bs{x}^{t,k-1}_i - \eta_l \nabla F_i(\bs{x}^{t,k-1}_i)$
        \ENDFOR
        \STATE Client uploads local updates $\bs{g}_i^t = \bs{x}^{t,0} - \bs{x}^{t, K}$ 
    \ENDFOR
    \STATE Server updates momentum $\bs{m}_i^{t}$ by Eq.~\eqref{eq:momentum_rule} \kern-18.5em\tikzmark{fedavg:begin} \kern+6.5em
    \STATE Server computes $\boldsymbol{\lambda}^t$ by Eq.~\eqref{eq:norm_minimize}
    \STATE Server computes $\bs{d}^t = \sum_{i=1}^N \boldsymbol{\lambda}^t_i \bs{m}_i^t$ \hfill \tikzmark{fedavg:end}
    \STATE Server updates $\bs{x}^{t+1} = \bs{x}^t - \eta_g \bs{d}^t$ 
    \drawCodeBox{fill=Blue}{fedavg:begin}{fedavg:end}
\ENDFOR
\end{algorithmic}
\end{algorithm}

\textbf{Adaptive aggregation via global norm minimization  }
In Eq.~\eqref{eq:norm_minimize}, we consider inputted $\{\bs{m}_i^t\}_{i\in[N]}$ as base vectors in $d$-dimensional linear space. This constrained minimization problem involves finding a minimum-norm point in the convex hull. Therefore, the attained estimates $\bs{d}$ is a vector from initial model parameters to this minimum-norm point. Besides, we particularly consider the non-convex optimization problem in FL, where the dimension of gradients can be millions due to the neural network scale (i.e., $d \gg N$). We use the Frank-Wolfe algorithm~\citep{jaggi2013revisiting} to solve it. Moreover, we argue that the norm of the global estimate is always non-zero during the whole FL training process, i.e., $\|\bs{d}^t\| > 0$ for all $t\in[T]$. In other words, these cases mean that all the vectors should be \textit{linearly independent}~\citep{greub2012linear}. Hence, we state that Algorithm~\ref{alg:gdm} does not fail by reaching some corner cases with $\|\bs{d}^t\| = 0$ in training neural network practices.

\textbf{Plugging AWARE in advanced federated algorithms  }
AWARE refers to the server-side aggregation module in Lines 10-12, Algorithm~\ref{alg:gdm}, which outputs a pseudo-gradient $\bs{d}^t$. It can enhance the generalization performance of existing federated algorithms by using the following extension:
\begin{proposition}[AWARE extension]\label{pro:projection} 
Given a federated optimization method updating the global model by $\bs{x}^{t+1} = \bs{x}^t - \eta_g \tilde{\bs{d}}^t$, where $\tilde{\bs{d}}^t$ is estimated by the method. We project the $\tilde{\bs{d}}^t$ to the direction of AWARE $\bs{d}^t$ by computing
$\bs{d}_{\text{proj}}^t = \frac{\langle \tilde{\bs{d}}^t, \bs{d}^t\rangle}{\langle \bs{d}^t, \bs{d}^t\rangle} \bs{d}^t,$
and then conduct $\bs{x}^{t+1} = \bs{x}^t - \eta_g \bs{d}_{\text{proj}}^t$.
\end{proposition}
This procedure only modifies the server-side gradient descent of applied algorithms to the direction that enhances generalization. We elaborate its insights in Section~\ref{sec:discussion} and evaluate it in Section~\ref{sec:exp}.

\section{Convergence Guarantees}\label{sec:guarantees}

In this section, we first provide deep insights into FedAvg's convergence using empirical properties. Then, by comparing with FedAvg, we theoretically demonstrate that using FedAWARE improves FL convergence. Then, we connect the results with existing theory to illustrate the generalization benefits.

\subsection{Common Theoretical Analysis of FedAvg}

We begin our analysis with the analytic assumption.
Conventional convergence analysis typically rely on common non-convex optimization assumptions~\citep{reddi2020adaptive, wang2022communication, acar2020federated} on local objectives $f_i(\bs{x}), i\in[N]$:
\begin{assumption}[Bounded dissimilarity]\label{asp:bgv}We assume the averaged global variance is bounded, i.e., $\sum_{i=1}^N\bs{\lambda}_i\mathbb{E}\left\|\nabla f_i(\bs{x})-\nabla f(\bs{x})\right\|^2 \leq \sigma_{g}^2$ for all $x \in \mathcal{X}$.
\end{assumption}
\begin{assumption}[Smoothness]\label{asp:smoothness}
    Each objective $f_i(\bs{x})$ for all $i\in[N]$ is $L$-smooth, inducing that for all $\forall \bs{x}, \bs{y}\in\mathbb{R}^d$, it holds $\|\nabla f_i(\bs{x}) - \nabla f_i(\bs{y})\| \leq L \|\bs{x} - \bs{y}\|$.
\end{assumption}
\begin{assumption}[Unbiasedness]\label{asp:unbiasedness}
    For each $i\in[N]$ and $\bs{x} \in \mathbb{R}^d$, we assume the access to an unbiased stochastic gradient $\nabla F_i(\bs{x}, \xi_i)$ of client's true gradient $\nabla f_i(\bs{x})$, i.e.,$\mathbb{E}_{\xi_i\sim\mathcal{D}_i}\left[\nabla F_i(\bs{x}, \xi_i)\right] =\nabla f_i(\bs{x})$.  The function $f_i$ have $\sigma_l$-bounded (local) variance i.e.,$\mathbb{E}_{\xi_i\sim\mathcal{D}_i}\left[\left\|\nabla F_i(\bs{x}, \xi_i)-\nabla f_i(\bs{x})\right\|^2\right] \leq \sigma_l^2$. 
\end{assumption}

\begin{table}[t]
\resizebox{\linewidth}{!}{
\begin{tabular}{lll}
\toprule
Terms & Definition & References \\ \midrule
grad. dissimilarity & $\mathbb{E}\|\nabla f_i(\bs{x})-\nabla f(\bs{x})\|^2 \leq \sigma_{g}^2$  &  \citep{woodworth2020minibatch}       \\
grad. diversity     & $\mathbb{E}\|\nabla f_i(\bs{x})\| \leq \gamma^2 \|\nabla f(\bs{x})\|$    & \citep{koloskova2020unified}         \\
general grad. diversity     & $\mathbb{E}\|\nabla f_i(\bs{x})\| \leq \gamma^2 \|\nabla f(\bs{x})\| + \sigma_g^2$   &  \citep{li2020federated}       \\
grad. norm     & $\mathbb{E}\|\nabla f_i(\bs{x})\| \leq \sigma_g^2$           &  \citep{li2019convergence}   \\\bottomrule
\end{tabular}}
\caption{Summary of data heterogeneity measures, adapted from Table 6~\citep{kairouz2021advances}.}\label{tab:summary}
\end{table}

The smoothness and unbiasedness are common assumptions in stochastic optimization analysis. The first assumption is specially made to analyze the convergence of FedAvg under heterogeneous clients. In Table~\ref{tab:summary}, we summarize the analogous heterogeneity measures used in the literature. We note that the existing convergence analysis of FedAvg has a similar dependency on $\sigma_g^2$, though $\sigma_g$'s definition is different. 

\textbf{Pessimistic convergence of FedAvg  }
These assumptions can be pessimistic in practice~\citep{wang2024on}.
Under the above assumptions, previous works derived an upper bound for the optimization error with non-convex objective functions, for example~\cite{jhunjhunwala2022fedvarp}:
\begin{theorem}\label{theorem:fedavg}
Under Assumptions~\ref{asp:bgv} to~\ref{asp:unbiasedness}, if FedAvg learning rates satisfy $\eta_l \leq 1/8LK, \eta \leq 1/24LK$, then the global gradient norm $\min_t \mathbb{E}\|\nabla f(\bs{x}^t)\|^2$ can be upper bounded by
\begin{equation}
\mathcal{O}\left(\frac{F^*}{\eta K T}\right)+\mathcal{O}\left(\eta \sigma_l^2 + \eta_l^2 K \sigma_l^2\right)+\mathcal{O}\left(\eta_l^2 K^2 \sigma_g^2\right),
\end{equation}
where $\eta = \eta_l\eta_g$, $f(\bs{x}^0) - f(\bs{x}^*) \leq F^*$ and $\bs{x}^* = \arg\min f(\bs{x})$.
\end{theorem}

Theorem~\ref{theorem:fedavg} shows that when the learning rates are fixed, and the communication round $T$ is limited, data heterogeneity always introduces an addition term $\mathcal{O}(\sigma_g^2)$ to the optimization error bound. Under heterogeneous clients, it indicates that FedAvg cannot outperform the simple mini-batch SGD~\citep{woodworth2020minibatch} or vanilla GD~\citep{khaled2020tighter}. Meanwhile, these upper bounds match a lower bound of FedAvg~\citep{glasgow2022sharp}, suggesting they are tight in the worst case. Therefore, we do not improve these bounds as they are already tight. Instead, we argue that the common heterogeneity measures are too pessimistic to cover the FedAvg practice, as the training dynamics of FedAvg do not always match the worst-case scenarios. We contend that existing convergence analyses overlook key properties in FedAvg's practical implementation, which are crucial for guiding the design and evaluation of federated algorithms.

\subsection{Convergence of FedAWARE under Client Consensus Dynamics}\label{sec:ccd}

We extend the advanced notion of heterogeneity, called \textit{client consensus hypothesis}~\citep{wang2024on}. This hypothesis states that \textit{the data heterogeneity does not have any negative impacts on the convergence of FedAvg if the norm of averaged local updates at \textbf{the global optimum} is close to zero.} This statement is proven under the strong convexity assumption. 
In contrast, we are interested in the non-convex optimization problem for providing accurate statements on the training of neural networks. To this end, we propose the \textit{client consensus dynamics} as shown below:
\begin{property}[Client Consensus Dynamics]\label{asp:client_consunsus}
The norm of $\bs{\lambda}$-averaged local updates (pseudo-gradient on server-size optimization in Eq.~\eqref{eq:framework})
$$
\rho^t(\bs{\lambda}) \triangleq \| \mathbb{E}[\boldsymbol{G}^t]\|^2 = \left\|\mathbb{E}\left[\sum_{i=1}^N \bs{\lambda}_i \bs{g}_i^t\right]\right\|^2
$$ 
decays along with federated optimization procedures, i.e., $\frac{1}{T}\sum_{t=1}^T \rho^t(\bs{\lambda}) \leq \mathcal{O}(T^{-c})$ and $c > 0$ in heterogeneous federated optimization.
\end{property}

We argue that the \textit{client consensus dynamics} decays with FL converges, as the global model typically converges to the points within the convex hull formed by heterogeneous local solutions. And, the local updates $\bs{g}_i^t$ reduces as it is closing to the local minimum along with the FL procedure. We provide empirical observations on this property in Figure~\ref{fig:observation}.
Moreover, we note that \textit{client consensus dynamics} is a weaker assumption than the \textit{client consensus hypothesis} as we do not assume the information at the global minimum. It allows us to analyze the FedAvg training dynamics without using pessimistic assumptions.
Now, we derive a new convergence analysis of FedAvg:
\begin{theorem}\label{theorem:fedavg_consensus} Under assumptions~\ref{asp:smoothness},~\ref{asp:unbiasedness} and property~\ref{asp:client_consunsus}, there exists static FedAvg learning rates $\eta_l, \eta_g$ such that the global gradient norm $\min_t \mathbb{E}\|\nabla f(\bs{x}^t)\|^2$ be upper bounded by
$$
\mathcal{O}(\frac{F^*}{\eta KT}) + \mathcal{O}(\eta\sigma_l^2 + \eta_l^2 K\sigma_l^2) +\mathcal{O}(
\eta \Psi),
$$
where 
$$
\Psi = \frac{1}{T}\sum_{t=0}^{T-1}\frac{\rho^t(\bs{\lambda})}{\eta_l^2K}.
$$
\end{theorem}
\textbf{Revisiting FedAvg convergence via client consensus dynamics  }
We can obtain the same results as Theorem~\ref{theorem:fedavg} if we explicitly derive the upper bound of term $\Psi$ using Assumption~\ref{asp:bgv}. For example, the local updates are upper bounded by $\bs{g}_i^t \leq K\eta_l^2(\sigma_l^2+6K\sigma_g^2)$ by this work~\citep{reddi2020adaptive}.
In other words, the dynamics term $\Psi$ absorbs the impacts of data heterogeneity and local gradient variances. The above theorem suggests that \textit{FedAvg typically converges with a better rate empirically than the rate described by Theorem~\ref{theorem:fedavg}}. This gap comes from the differences between pessimistic assumptions and empirical client consensus dynamics.
Moreover, using an optimized learning rate (e.g., Lemma~\ref{lemma:constant_stepsize}) in Appendix~\ref{app:lemmas}, the above upper bound can achieve a dominated rate of $\mathcal{O}(\Psi/\sqrt{T})$. It indicates that \textit{the data heterogeneity has less impact on the training procedure (i.e., $\{\bs{x}^t\}_{t=0}^{T-1}$) if the attained client consensus dynamics $\Psi$ is small}.
Therefore, a better aggregation strategy that reduces client consensus dynamics is a promising way to enhance FL under heterogeneity scenarios.

However, existing convergence analyses of FedAvg typically use uniform weights and pessimistic heterogeneity measures. They omitted the power of global aggregation in FedAvg. And, how the aggregation weights $\bs{\lambda}$ work in the convergence of FedAvg remains ambiguous. The commonly used static weight $\bs{\lambda}$ is typically related to the global objective. The previous discussion motivates us to investigate the convergence of FedAvg when the global objective weight $\bs{\lambda}$ and the applied aggregation weight $\tilde{\bs{\lambda}}$ are not identical. To describe the power of aggregation weights, we present the following Corollary~\ref{cor:adaptive_fedavg}: 
\begin{corollary}\label{cor:adaptive_fedavg}
Suppose FL defines its global objective with a static weight $\bs{\lambda}$ while aggregating local updates with (possibly adaptive) weights $\tilde{\bs{\lambda}}$ such that $\tilde{\bs{G}}^t = \sum_{i=1}^{N}\tilde{\bs{\lambda}}_i \bs{g}_i^t$. If FedAvg uses $\tilde{\bs{G}}^t$ for global gradient descent instead of $\bs{G}^t$ in Eq.~\eqref{eq:framework}. Then, the global gradient norm $\min_t \mathbb{E}\|\nabla f(\bs{x}^t)\|^2$ can be upper bounded by
$$
\mathcal{O}(\frac{F^*}{\eta KT}) + \mathcal{O}(\eta\sigma_l^2 + \eta_l^2 K\sigma_l^2) +\mathcal{O}(
\eta \tilde{\Psi} ),
$$
where 
$$
\tilde{\Psi} = \frac{1}{T}\sum_{t=0}^{T-1} \chi_{\bs{\lambda}\|\tilde{\bs{\lambda}}}^2  \cdot \frac{\rho^t(\tilde{\bs{\lambda}})}{\eta_l^2K}, \; \chi_{\bs{\lambda}\|\tilde{\bs{\lambda}}}^2 = \sum_{i=1}^N \frac{(\bs{\lambda}_i - \tilde{\bs{\lambda}}_i)^2}{\bs{\lambda}_i}.
$$
\end{corollary}

Compared with Theorem~\ref{theorem:fedavg_consensus}, the above corollary indicates that \textit{adptive aggregation strategy accelerates the FL convergence if it induces lower client consensus dynamics $\tilde{\Psi}$ on a training trajectory of $\{\bs{x}^t\}_{t=0}^{T-1}$}. It shows that a key step in enhancing FedAvg is to design the pseudo-gradient for global updates and achieve a promising client consensus, i.e., minimize the norm of the applied pseudo-gradient. 

\textbf{Convergence of FedAWARE  }
The convergence benefits of FedAWARE follow the afore-discussed insights. 
In the methodology, we take the moving-averaged local updates $\bs{m}_i^t$ as the approximation of local updates $\bs{g}_i^t$. The solution of Eq.\eqref{eq:norm_minimize} is to minimize client consensus for the current communication round greedily. When $\alpha=1$ and full client participation, the above Corollary matches the convergence of FedAWARE, ensuring it converges faster than vanilla FedAvg in heterogeneous environments. Further empirical study demonstrates surprisingly superior empirical convergence of FedAWARE compared to more advanced federated algorithms.
In the Appendix, we present the convergence analysis of FedAWARE with partial client participation, which is analogous to the corollary. The results indicate that a proper selection of $\alpha$ can further mitigate the impacts of local gradient variances. Further ablation studies provide evidence of its efficiency.

\subsection{Why does FedAWARE Generalize Better?}\label{sec:discussion}

FedAWARE achieves better generalization performance by potentially enlarging the gradient diversity over the FL training process.
Gradient diversity is first introduced in data-centralized distributed learning~\citep{yin2018gradient}, which quantifies the degree to which individual gradients diverge from each other. It is proved that \textit{distributed mini-batch SGD is amenable to better speedups and generalization if the problems attain high gradient diversity}. Moreover, the common heterogeneity measures are its extensions in federated optimization analysis~\citep{haddadpour2019convergence, li2020federated}. However, vanilla gradient diversity typically serves as a theoretical analysis tool, lacking empirical guidance for federated algorithm design.
To resolve this drawback, we replace the local first-order gradient $\nabla f_i(\bs{x}^t)$ with local updates $\bs{g}_i^t$. 
Formally, we present an empirical  measurement termed as \textit{Local Update Diversity} (LUD):
\begin{definition}[Local Update Diversity]\label{def:diversity} For any round $t\in[T]$ of the FL training process, we define the local update diversity by
\begin{equation}\label{eq:diversity}
\delta_D^t \triangleq \sqrt{\frac{\sum_{i=1}^N \bs{\lambda}_i \mathbb{E}\|\bs{g}_i^t\|^2}{\mathbb{E}\|\bs{G}^t\|^2}} = \sqrt{\frac{\sum_{i=1}^N \bs{\lambda}_i \mathbb{E}\|\bs{g}_i^t\|^2}{\mathbb{V}(\bs{G}^t) + \rho^t(\bs{\lambda})}}.
\end{equation}
\end{definition}
LUD quantifies the degree of local updates $\bs{g}_i^t$ diverse from global pseudo-gradient $\bs{G}^t$. It also reflects the ratio of convergence rates between local and global update norms. In non-convex FL, we propose to track the LUD dynamics, that is, the evolution of LUD values in an FL training process. Here, we discuss deep insights into the LUD dynamics.

\textbf{FL is a process of enlarging LUD  } 
Intuitively, when the model is far from the global solution, most of the gradients point in a similar direction, which means that gradient diversity is initially small. When the model approaches the global solution, the LUD starts to grow because of the heterogeneity at the local optimums. From another perspective, this intuition also matches the client consensus dynamics, as shown in Eq.~\eqref{eq:diversity}. Straightforwardly, the client's consensus $\rho^t(\bs{\lambda})$ typically decreases faster than local updates during training due to data heterogeneity. Therefore, the empirical LUD values are increasing during training. 

\begin{figure}[t]
\centering
\includegraphics[width=0.9\linewidth]{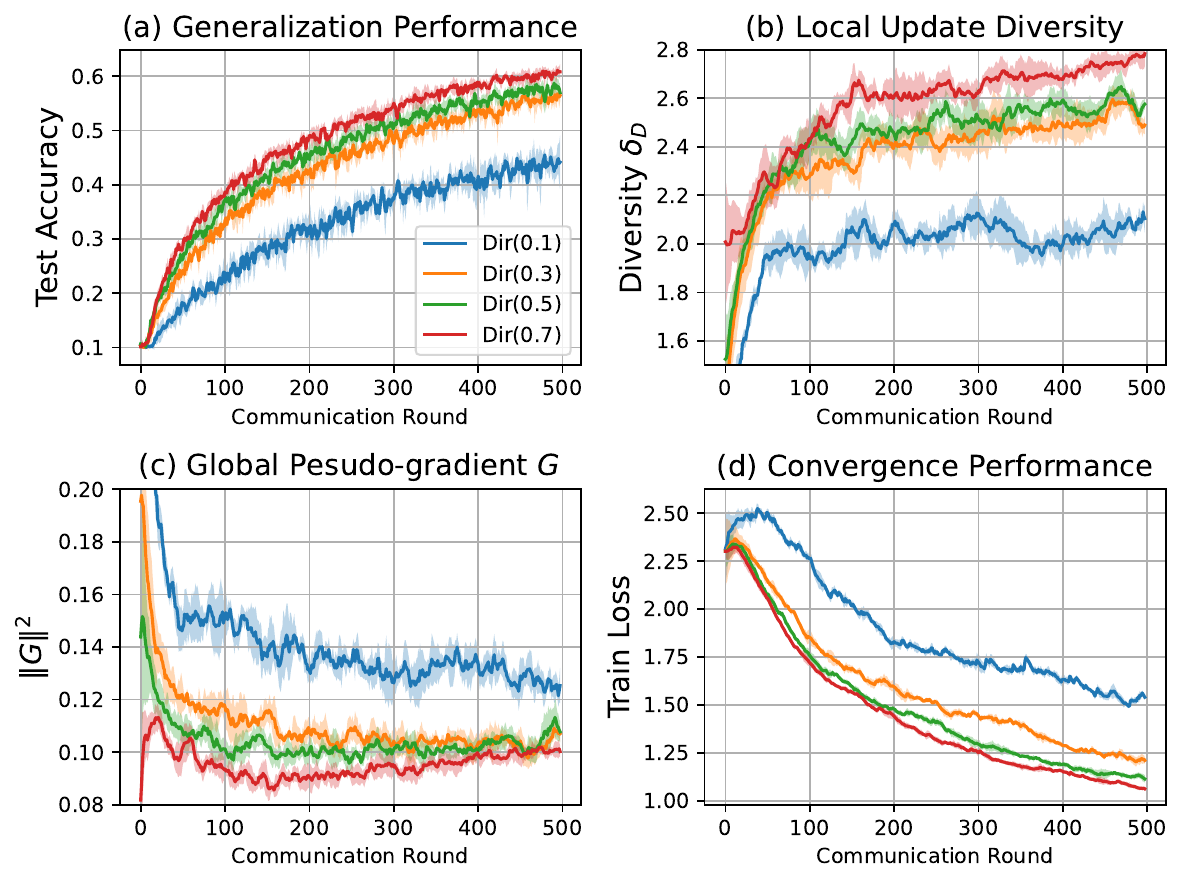}
\caption{Training dynamics of FedAvg on Non-IID partitioned CIFAR-10 task. We use Dirichlet distribution to allocate clients' data as described in Section~\ref{sec:exp}. Dir(0.1) indicates the most heterogeneous FL setting. }\label{fig:observation}
\end{figure}

To illustrate this statement, we observe the training dynamics of FedAvg on different degrees of Non-IID partitioned CIFAR-10 experiments in Figure~\ref{fig:observation}. Note that different curves indicate the different levels of data heterogeneity that impact FedAvg convergence. We observe that the LUD values increase over the communication round in all heterogeneity settings. Moreover, Figure~\ref{fig:observation}(b) demonstrates that FedAvg attains higher LUD when the heterogeneity level is low. It obtains faster convergence curves, as shown in Figure~\ref{fig:observation}(d). Besides, Figure~\ref{fig:observation}(c) provides empirical evidence for the Assumption~\ref{asp:client_consunsus}, showing that the norm of global pseudo-gradient decays during FL training.

\textbf{FedAWARE further enlarges LUD dynamics }
Gradient diversity\cite{yin2018gradient} has shown that distributed mini-batch SGD gains its generalization ability with larger gradient diversity. As FedAvg can be interpreted as a perturbed distributed SGD, this conclusion should apply to FedAvg and its variants. For example, Figure~\ref{fig:observation} demonstrates that FedAvg archives higher local update diversity (Figure\ref{fig:observation} (b)) and tends to obtain better test accuracy (Figure~\ref{fig:observation} (a)).  
In the experimental evaluation, we empirically illustrate that FedAWARE further enlarges the LUD dynamics during training, thus obtaining better generalization performance.
Future extension of algorithmic stability~\cite{hardt2016train, lei2020fine} on FedAWARE may provide deeper theoretical insights into its generalization ability.

\newcommand{\ad}[1]{\textsubscript{\textcolor{red}{$+$#1}}}
\newcommand{\dc}[1]{\textsubscript{\textcolor{black}{$-$#1}}}
\begin{table*}[t]
\centering
\caption{Evaluation of CIFAR-10, CIFAR-100 and AGNews tasks.
In the table, Raw denotes the original version of all algorithms. And, $\times$AWARE denotes using for FedAWARE extension in Proposition~\ref{pro:projection} on corresponding methods. And, e-LUDD denotes e-LUD dynamics (i.e., $\sum_{t=0}^{T-1}\tilde{\delta}_D^t/T$) during training. 
The numbers in red indicate the improvements after applying $\times$AWARE. 
We run $T$ communication rounds for stable test accuracy. We report the mean test accuracy of the last 10\% communication rounds, indicating generalization stability.}\label{tab:main_results}
\resizebox{\linewidth}{!}{
\begin{tabular}{lcc|cc|cc|cc|cc|cc}
\toprule
\multicolumn{1}{c}{\multirow{3}{*}{Settings}} & \multicolumn{4}{c}{CIFAR-10, $T=500$}                                  & \multicolumn{4}{c}{CIFAR-100, $T=600$}                                 & \multicolumn{4}{c}{AGNews, $T=300$}                                   \\ \cmidrule(r){2-5} \cmidrule(r){6-9} \cmidrule(r){10-13}
\multicolumn{1}{c}{}                           & \multicolumn{2}{c}{Raw} & \multicolumn{2}{c}{$\times$AWARE} & \multicolumn{2}{c}{Raw } & \multicolumn{2}{c}{$\times$AWARE} & \multicolumn{2}{c}{Raw} & \multicolumn{2}{c}{$\times$AWARE}  \\ \cmidrule(r){2-3} \cmidrule(r){4-5} \cmidrule(r){6-7} \cmidrule(r){8-9} \cmidrule(r){10-11} \cmidrule(r){12-13}
\multicolumn{1}{c}{}                           & \multicolumn{1}{c}{Acc.} & e-LUDD & \multicolumn{1}{c}{Acc.} & e-LUDD & \multicolumn{1}{c}{Acc.} & e-LUDD & \multicolumn{1}{c}{Acc.} & e-LUDD & \multicolumn{1}{c}{Acc.} & e-LUDD & \multicolumn{1}{c}{Acc.} & e-LUDD \\ \midrule
FedAvg     &     42.78           &  2.58 & 49.25\ad{6.47}  &   3.08\ad{0.50}       &     28.69     &   2.94    &  32.72\ad{4.03}      &  3.61\ad{0.67}      &  79.27      &  1.25   & 80.21\ad{0.94}    &   1.98\ad{0.73}      \\
FedAvgM    &     50.07           &  2.17 & \textbf{59.58}\ad{9.51}  &   2.92\ad{0.75}  &     31.04     &  2.72    &   40.66\ad{9.62}   &   3.37\ad{0.65}        &  81.85      &  1.38   &  83.15\ad{1.30}    &   1.85\ad{0.47}           \\
FedCM      &     47.84           &  2.99 & 49.13\ad{1.29}  &   3.09\ad{0.10}  &     30.62     &  2.94           &   33.09\ad{2.04}    &   3.63\ad{0.69}  &  80.79      &  1.26   &   80.24\dc{0.55}      &   1.76\ad{0.50}        \\
FedDyn     &     54.87           &   -   &    -             &     -   &      \textbf{39.04} &    -     &      -      &       -      &  \underline{84.47}      &  -    &         -     &     -       \\
FedProx    &     46.33           &  2.81 & 49.40\ad{3.07}  &  2.87\ad{0.06}   &  27.20   &     2.95        &  32.68\ad{5.48}     &   3.62\ad{0.67}    &  76.81      &  1.15    &    80.13\ad{3.33}      &    1.83\ad{0.68}           \\
FedYogi    &     54.62           &  2.47 & 54.92\ad{0.30}   & 2.66\ad{0.19} &     37.12     &     2.61        &   \underline{44.02}\ad{6.90}     &   3.27\ad{0.66}     &  83.15      & 1.43     &  \underline{85.89}\ad{2.74}     &     2.01\ad{0.58}          \\
FedAMS     &     \underline{57.09}           &  2.46 &   \underline{58.45}\ad{1.36}      &  2.67\ad{0.21}   &     38.19     &     2.42        &   \textbf{45.20}\ad{7.01}     &   3.13\ad{0.71}     &  83.33      &  1.34    &   \textbf{86.08}\ad{2.75}             &   1.92\ad{0.58}        \\
FedAWARE   &    \textbf{59.78}   &  2.90 &    -             &    -    &     \underline{38.77}     &     2.75        &        -    &      -       &  \textbf{87.70}      &  1.73    &          -    &    -        \\ \bottomrule
\end{tabular}}
\end{table*}

\begin{figure*}[t]
\centering
\includegraphics[width=0.9\linewidth]{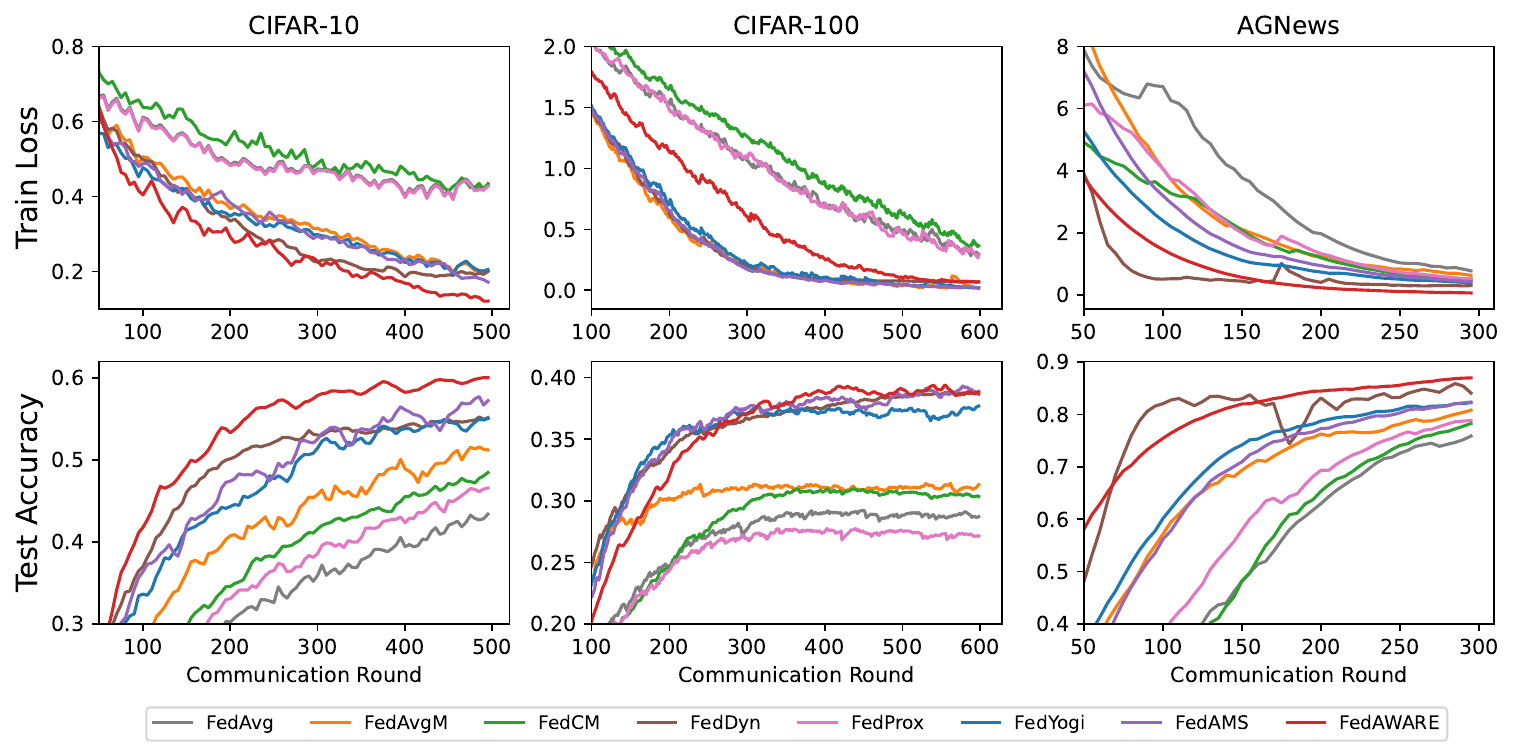}
\caption{Training dynamics of raw algorithms. The training loss indicates the convergence speed on training datasets, while the test accuracy indicates the generalization stability against heterogeneous clients.}\label{fig:main_results}
\end{figure*}

\textbf{Discussion: connection with client coherence  } 
Client coherence~\citep{Chatterjee20, li2023revisiting}, defined as the sum of cosine similarities between the local updates of distinct clients, is also used to quantify heterogeneity across clients. The global pseudo-gradient norm (LUD denominator) is related to client coherence, incorporating additional information about the local update norms. A recent study~\citep{li2023revisiting} states: \textit{Gradient diversity argues that higher similarities between workers’ gradients degrade performance in distributed mini-batch SGD, while gradient coherence claims that higher similarities between sample gradients enhance generalization.} However, these concepts are not contradictory, as they pertain to different training stages. In FL, client coherence is relevant only before a critical point where overall coherence becomes positive~\citep{li2023revisiting}, typically during the early training rounds. By contrast, LUD dynamics characterize the entire FL training process, with larger overall LUD dynamics consistently associated with improved generalization, as demonstrated in our experiments. We further elaborate on these points in Appendix~\ref{app:exp_comparison}.

\textbf{Discussion: LUD quantifies heterogeneity better }
Common data heterogeneity measures/assumptions in Table~\ref{tab:summary} typically build on the local and global first-order gradients. In FL, these common measures cannot be computed in practice due to privacy risk~\citep{zhu2019deep} or communication efficiency~\citep{mcmahan2017communication}. And, client coherence only measures the heterogeneity in local update directions.
Hence, they may overlook the effects of unbalanced local client updates. For example, local updates typically consist of cumulative stochastic mini-batch gradients computed over multiple epochs on the local dataset. When the number of local data samples varies significantly across clients, LUD also reflects the imbalance in local updates. Therefore, the LUD metric provides more comprehensive information than gradient diversity in FL systems. We propose tracking the evolution of $\delta_D^t$ to quantify the impact of heterogeneity and assess the quality of FL convergence.

\section{Experimental Evaluation}\label{sec:exp}

\textbf{Baselines  } 
We compare advanced FL algorithms and basic baselines related to the methodologies of FedAWARE. Our baselines include standard baseline FedAvg~\citep{mcmahan2017communication}, FedAvgM~\citep{hsu2019measuring}, and FedCM~\citep{wang2022communication} for the static aggregation and momentum-based algorithms. We also compare with local regularization-based algorithms FedProx~\citep{li2020federated} and FedDyn~\citep{acar2020federated}. We compare the adaptive federated optimization algorithms with FedYogi~\citep{reddi2020adaptive} and FedAMS~\citep{wang2022communication}.

\textbf{Settings  } 
We evaluate all algorithms on three setups: (1) train 5-layer CNN on Non-IID partitioned CIFAR-10 dataset. (2) train Resnet-18 (group norm)~\citep{he2016deep} on Non-IID partitioned CIFAR-100 dataset. (3) fine-tune pretrained GPT2 model Pythia-70M~\citep{biderman2023pythia} on Non-IID partitioned AGNews dataset~\citep{Zhang2015CharacterlevelCN}. 
CIFAR-10 and CIFAR-100 are image classification tasks with 50,000 train data samples and 10,000 test data samples. AGNews is a collection of more than 1 million news articles. It is a four-label text classification task with 120,000 train samples and 7,600 test samples.
For all datasets, we conduct Non-IID data partitioning following the latent Dirichlet allocation over labels~\citep{hsu2019measuring} with parameters 0.1 into $N=100$ clients, indicating an extreme data heterogeneity. The visualization of datasets is provided in the Appendix.
For the training setup, we set learning rate $\eta_l=0.01$ (CIFAR) or $\eta_l=0.0001$ (AGNews) for local training parameters, batch size $64$, and local epoch $3$ for all clients. For each communication round, we randomly select 10 clients. Then, we grid search global learning rate $\eta_g$ and the algorithm-specific hyperparameters for all algorithms as elaborated in the Appendix. All results are the mean values of three independent runs over different random seeds.

\textbf{Metrics  }
We evaluate the test accuracy of algorithms on RAW test datasets and observe the training loss dynamics to compare the convergence speed. Moreover, we also observe the LUD dynamics of FL. However, we cannot accurately compute the LUD in practice due to the partial participation of clients. Instead, we define a metric called an empirical LUD (e-LUD):
$$
    \tilde{\delta}_D^t \triangleq \sqrt{\frac{\frac{1}{|S^t|}\sum_{i\in S^t} \|\bs{g}_i^t\|^2}{\| \frac{1}{|S^t|}\sum_{i\in S^t} \bs{g}_i^t \|^2}},
$$
where $S^t$ is the selected client set at the $t$-th round and $\bs{g}_i^t$ is the uploaded local updates from selected clients. For a fair comparison, the e-LUD of all algorithms is computed the same. Their values indicate the status of global solutions and their relation to local minimums.

\begin{figure}[t]
\centering
\includegraphics[width=\linewidth]{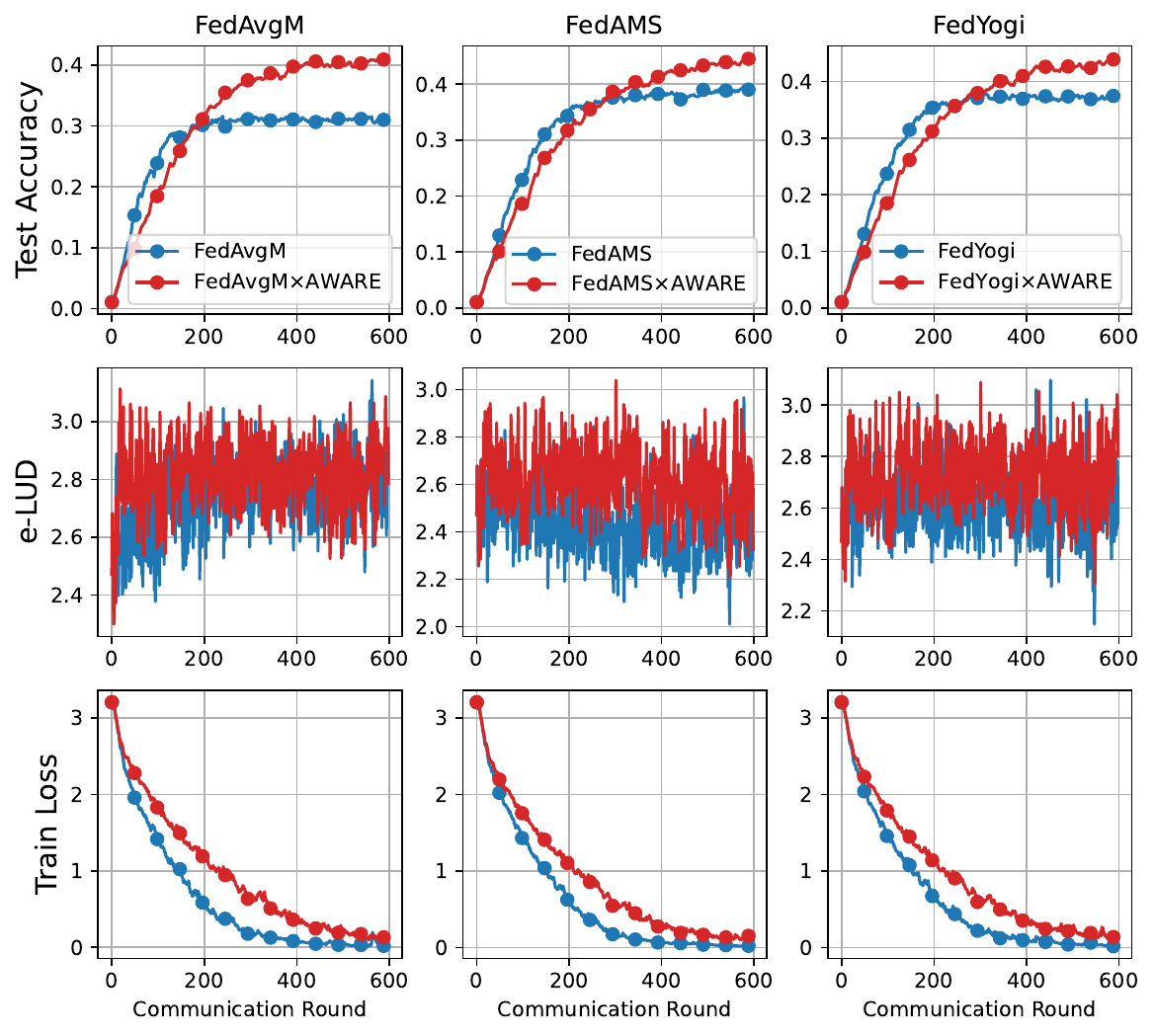}
\vspace{-3mm}
\caption{Training dynamics of FedAWARE extension on CIFAR-100 setting. The results of CIFAR-10 and AGNews are presented in the Appendix.}\label{fig:cifar100-projection}
\vspace{-3mm}
\end{figure}

\textbf{Vanilla FedAWARE converges faster and generalizes better  }
In Figure~\ref{fig:main_results}, we examine the convergence performance of algorithms. The results highlight FedAWARE's superior performance. On CIFAR-10 and AGNews tasks, we surprisingly observed that FedAWARE converges faster than most baselines and even outperforms adaptive federated optimization methods (FedYogi and FedAMS). On the CIFAR-100 task, despite FedAWARE converging slightly slower than the SOTA methods at early communication rounds, it still reaches a good solution with identical train loss. Meanwhile, the test accuracy of FedAWARE also implements marginal improvement to FedAvgM, FedYogi, and FedAMS methods, as shown in Table~\ref{tab:main_results}.  
Given the empirical results, we argue that FedAWARE, a novel variant of FedAvg, is more efficient for handling heterogeneous local updates. Besides noting that the key performance improvement of FedAWARE is obtained from the adaptive weighted aggregation strategies, the empirical results also provide solid support for our Corollary~\ref{cor:adaptive_fedavg}.

\begin{figure}[t]
\centering
\includegraphics[width=\linewidth]{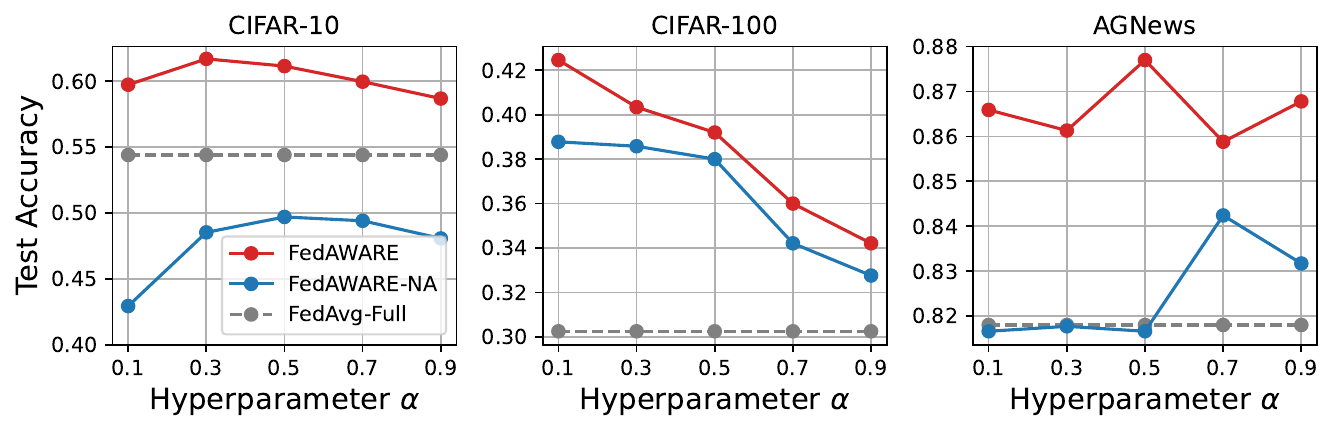}
\vspace{-3mm}
\caption{Ablation study. "FedAWARE-NA" means we only use moving-averaged local updates without adaptive aggregation. "FedAvg-Full" is the result of vanilla FedAvg with full client participation.}\label{fig:ablation_study}
\vspace{-3mm}
\end{figure}

\textbf{Enlarging e-LUD dynamics of algorithms obtain better generalization performance  }
In Table~\ref{tab:main_results}, we apply the FedAWARE extension (Proposition \ref{pro:projection}) on all baseline algorithms, excluding FedDyn, which does not follow the global gradient descent rule. Compared with the raw results of algorithms, the FedAWARE module enlarges the e-LUDD of ALL algorithms. Meanwhile, the test accuracy of most algorithms is significantly improved. Surprisingly, the FedAMS$\times$AWARE on CIFAR100 implements superior accuracy than all raw algorithms. The results provide empirical evidence on the relation between LUD and generalization, that is, \textit{federated learning is amenable to better generalization if the training procedure attains high LUD}. We believe this finding can enlighten future techniques for enhancing FL generalization.

\textbf{FedAWARE extension may slow convergence slightly in exchange for generalization boost  }
In Table~\ref{tab:main_results}, we observe that FedAWARE greatly enhances FedAvgM, FedYogi, and FedAMS. To provide deeper insights. we further examine the training dynamics in CIFAR-100 task after applying the FedAWARE extension, as shown in Figure~\ref{fig:cifar100-projection}. The convergence speed of applied algorithms has decreased slightly, and we observed that the test accuracy curves reached a higher final accuracy. Moreover, additional results on the CIFAR-10 and AGNews tasks in the Appendix show that the FedAWARE extension mitigates the test accuracy spikes problem, indicating stabilized generalization performance. The results demonstrate that enlarge e-LUD dynamics of algorithms also stabilize the generalization of a training procedure.

\textbf{Ablation Study} 
In Figure~\ref{fig:ablation_study}, we present the ablation study to evaluate the components of FedAWARE. We run FedAWARE with different moving-average parameters $\alpha=\{0.1, 0.3, 0.5, 0.7, 0.9\}$ to evaluate its sensitivity. The results show that the proper selection of $\alpha$ can enhance the generalization performance, while the best selection varies on different tasks. 
We also conducted the same experiments without adaptive aggregation; FedAWARE's results are better than all FedAWARE-NA results. Meanwhile, it is worth noting that FedAWARE's e-LUDD is also higher than FedAWARE-NA's. Compared with FedAvg-Full, moving-averaged local updates only provide marginal improvements on CIFAR-100 and AGNews tasks. This indicates the critical ability of adaptive aggregation.

\textbf{Limitation and feasible solutions  }
Running Algorithm~\ref{alg:gdm} requires the server to store the local momentum. We argue that the FL server has sufficient storage and computing resources. Besides, we argue this memory consumption is worth using FedAWARE to enhance other algorithms. This concern can be alleviated by the following options to save the storage: \emph{1. Computing~\eqref{eq:norm_minimize} with the last few layers of a neural network. }This is because numerous studies~\citep{xu2020neural, kirichenko2022last, burns2023weak} have shown that the last layers contain crucial network information. Hence, we can trade off the memory storage and quality of the aggregation. \emph{2. Adopting client clustering techniques}~\citep{sattler2020clustered, li2021federated, long2023multi}. We can cluster clients respecting their similarity and save the momentum of cluster-averaged gradients. The server only costs the storage proportional to the number of client clusters.

We have empirically demonstrated that enlarging the LUD dynamics of baseline algorithms by incorporating AWARE can improve generalization performance. However, the causal relation between LUD and model performance remains inconclusive due to a gap between the practical use of e-LUD and the theoretical LUD measure. For example, methods with the highest accuracy do not necessarily have the highest e-LUDD values and may even have relatively low values (e.g., FedAMS in "CIFAR-100 xAWARE") in Table~\ref{tab:main_results}. Hence, investigating correlations between absolute heterogeneity measures and model performance may provide valuable insights for future research. Additionally, we propose that the theoretical relationship between LUD and generalization performance could be established through algorithmic stability analyses~\citep{lei2020fine} or our recent framework for analyzing excess risk dynamics~\citep{zeng2024understanding}, which we leave for future exploration.

\section{Conclusion}

In this paper, we revisit the power of FedAvg aggregation in theory and practice. 
Our results further enrich the understanding of FedAvg and its variants by highlighting client consensus dynamics in convergence analysis and local update diversity in generalization. 
Based on our findings, we derive a simple and effective FedAvg variant, FedAWARE, which minimizes the norm of global aggregation results for convergence and generalization performance. 
Importantly, given the strong relationship between LUD dynamics and FL generalization illustrated by intensive empirical results, we believe LUD can be a critical measurement for designing and evaluating future federated algorithms.

\bibliography{main}
\bibliographystyle{apalike}

\onecolumn
\appendix
\startcontents[sections]
\printcontents[sections]{l}{1}{\setcounter{tocdepth}{5}}

\section{Convergence Guarantees}

\subsection{Useful Lemmas}\label{app:lemmas}

\begin{lemma}[Bounded local updates~\citep{reddi2020adaptive}]\label{lemma:local_drift} Let Assumption~\ref{asp:unbiasedness}~\ref{asp:bgv} hold. For all client $i\in[N]$ with arbitrary local iteration steps $k \in [K]$ and $\eta_l \leq \frac{1}{8KL}$, the local updates can be bounded as follows,
$$
\mathbb{E}\left\|\bs{x}_i^{t,k}-\bs{x}^t\right\|^2 \leq 5K\eta_l^2 (\sigma_l^2 + 6K\sigma_g^2 + 6K\left\| \nabla f\left(\bs{x}^t\right) \right\|^2).
$$
\end{lemma}

\begin{lemma}[Tuning the stepsize~\citep{koloskova2020unified}]\label{lemma:constant_stepsize} For any parameters $r_0 \geq 0, b \geq 0, e \geq 0, d \geq 0$ there exists constant stepsize $\eta \leq \frac{1}{d}$ such that
$$
\Psi_T:=\frac{r_0}{\eta T}+b \eta+e \eta^2 \leq 2\left(\frac{b r_0}{T}\right)^{\frac{1}{2}}+2 e^{1 / 3}\left(\frac{r_0}{T}\right)^{\frac{2}{3}}+\frac{d r_0}{T}.       
$$
\end{lemma}

\subsection{Proof of Theorem~\ref{theorem:fedavg_consensus}}

Here, we prove the convergence rate of FedAvg:
$$
\begin{aligned}
& \text{\textbf{Client: }}\bs{g}_i^{t} = \bs{x}_i^{t, K} - \bs{x}_i^{t, 0} = \eta_l \sum_{k=0}^{K-1} \nabla F_i(\bs{x}_i^{t, k});\\
& \text{\textbf{Server: }}\bs{x}^{t+1} = \bs{x}^{t}-\eta_g\sum_{i=1}^N \bs{\lambda}_i \bs{g}_i^{t} = \bs{x}^{t}- \eta_g \boldsymbol{G}^t.
\end{aligned}
$$
Specifically, we analyze the update rule on the server-side gradient descent
$$
\bs{x}^{t+1} = \bs{x}^{t} - \eta_g \bs{G}^t.
$$
For ease of writing, we rewrite the above equations as follows
$$
\bs{x}^{t+1} = \bs{x}^{t} - \tilde{\eta} \hat{\bs{G}}^t,
$$
where $\tilde{\eta} = K \eta_l\eta_g$ and $\hat{\bs{G}}^t = \bs{G}^t / \eta_l K = \frac{1}{NK}\sum_{i=1}^{N} \sum_{k=0}^{K-1} \nabla F_i(\bs{x}_i^{t, k})$. 

Using the smoothness, we have
$$
\begin{aligned}
f\left(\bs{x}^{t+1}\right) & = f\left(\bs{x}^{t}-\tilde{\eta} \hat{\bs{G}}^t\right) \leq f(\bs{x}^t) - \tilde{\eta} \left\langle\nabla f(\bs{x}^t), \hat{\bs{G}}^t\right\rangle + \frac{L}{2}\tilde{\eta}^2\left\|\hat{\bs{G}}^t\right\|^2 \\
& \leq f(\bs{x}^t) - \tilde{\eta} \left\langle\nabla f(\bs{x}^t), \hat{\bs{G}}^t - \nabla f(\bs{x}^t) + \nabla f(\bs{x}^t) \right\rangle + \frac{L}{2}\tilde{\eta}^2\left\|\hat{\bs{G}}^t\right\|^2 \\ 
& \leq f(\bs{x}^t) - \tilde{\eta} \|\nabla f(\bs{x}^t)\|^2 + \tilde{\eta} \left\langle\nabla f(\bs{x}^t), \nabla f(\bs{x}^t) - \hat{\bs{G}}^t \right\rangle + \frac{L}{2}\tilde{\eta}^2\left\|\hat{\bs{G}}^t\right\|^2 \\ 
& \leq f(\bs{x}^t) - \frac{\tilde{\eta}}{2} \|\nabla f(\bs{x}^t)\|^2 + \frac{\tilde{\eta}}{2} \|\nabla f(\bs{x}^t) - \hat{\bs{G}}^t\|^2 + \frac{L}{2}\tilde{\eta}^2\left\|\hat{\bs{G}}^t\right\|^2.
\end{aligned}
$$

Taking full expectation over randomness at time step $t$ on both sides, we have
\begin{equation}\label{eq:fedavg_descent}
\begin{aligned}
\mathbb{E}\left[f(\bs{x}^{t+1})\right] - f(\bs{x}^{t}) & \leq - \frac{\tilde{\eta}}{2} \mathbb{E}\|\nabla f(\bs{x}^t)\|^2 + \frac{\tilde{\eta}}{2}\mathbb{E}\|\nabla f(\bs{x}^t) - \hat{\bs{G}}^t\|^2 + \frac{L}{2} \tilde{\eta}^2 \mathbb{E}[\|\hat{\bs{G}}^t\|^2] 
\end{aligned}
\end{equation}

Then, we investigate
$$
\begin{aligned}
\mathbb{E}\|\nabla f(\bs{x}^t) - \hat{\bs{G}}^t\|^2 & = \mathbb{E}\|\sum_{i=1}^N \bs{\lambda}_i (\nabla f_i\left(\bs{x}^t\right) - \frac{1}{\eta_l K}\bs{g}_i^t)\|^2 \leq \sum_{i=1}^N \bs{\lambda}_i \mathbb{E}\|\nabla f_i\left(\bs{x}^t\right) - \frac{1}{\eta_l K} \bs{g}_i^t\|^2 \\
& = \sum_{i=1}^N \bs{\lambda}_i \mathbb{E}\left\| \left(\nabla f_i\left(\bs{x}^t\right) - \frac{1}{K} \sum_{k=0}^{K-1} \nabla F_i(\bs{x}_i^{t, k})\right)\right\|^2 \\
& = \sum_{i=1}^N \bs{\lambda}_i \mathbb{E}\left\| \left(\frac{1}{K}\sum_{k=0}^{K-1}(\nabla f_i\left(\bs{x}^t\right) - \nabla F_i(\bs{x}_i^{t, k}))\right)\right\|^2  \\
& \leq \frac{1}{K^2} \sum_{i=1}^N \bs{\lambda}_i \mathbb{E}\left\| \left(\sum_{k=0}^{K-1}(\nabla f_i\left(\bs{x}^t\right) \pm \nabla f_i(\bs{x}^{t,k}) - \nabla F_i(\bs{x}_i^{t, k}))\right)\right\|^2 \\
& \leq 2\frac{1}{K^2}\sum_{i=1}^N \bs{\lambda}_i\mathbb{E}\left\|\sum_{k=0}^{K-1}(\nabla f_i\left(\bs{x}^t\right) - \nabla f_i(\bs{x}^{t,k}))\right\|^2 + 2\frac{1}{K^2}\sum_{i=1}^N \bs{\lambda}_i \mathbb{E}\left\|\sum_{k=0}^{K-1} (\nabla f_i(\bs{x}^{t,k}) - \nabla F_i(\bs{x}_i^{t, k}))\right\|^2 \\
& \leq 2\frac{1}{K^2}\sum_{i=1}^N \bs{\lambda}_i\mathbb{E}\left\|\sum_{k=0}^{K-1}(\nabla f_i\left(\bs{x}^t\right) - \nabla f_i(\bs{x}^{t,k}))\right\|^2 + 2\sigma_l^2/K \\
 & \leq 2\frac{1}{K^2} L^2 \sum_{i=1}^N \bs{\lambda}_i K\sum_{k=0}^{K-1} \mathbb{E}\left\|\bs{x}^t - \bs{x}^{t, k-1}_i \right\|^2 + 2c\eta_l^2L^2K \sigma_l^2 \quad\quad\quad \triangleright \text{Letting  } \frac
 {1}{\sqrt{c}KL}\leq \eta_l \\
& \leq 2 L^2 \sum_{i=1}^N \bs{\lambda}_i\mathbb{E}\left\| \bs{g}_i^t\right\|^2 + 2c\eta_l^2L^2K \sigma_l^2 \\
\end{aligned}
$$

This work mainly focuses on the benefits of using adaptive aggregation. To simplify the analysis, we make a mild assumption on the relation between the local update and the global pseudo-gradient norm:
\begin{assumption}\label{asp:norm_relation}
We assume there is a constant $\gamma$ makes the local update $\bs{g}_i^t$ and applied global pseudo-gradient $\bs{G}^t$ satisfies that
$$
\mathbb{E}\|\bs{g}_i^t\|^2 \leq \gamma \mathbb{E}\|\bs{G}^t\|^2,
$$
$\forall t\in[T], i\in[N], \bs{\lambda} \in \Delta_N$, where $\Delta_N$ is the N-dimensional simplex. Please note the $\gamma$ value will not exceed a minor constant in practice.
\end{assumption}

Using the above assumption, we obtain
\begin{equation}\label{tmp:t_1}
\begin{aligned}
\mathbb{E}\|\nabla f(\bs{x}^t) - \hat{\bs{G}}^t\|^2 & \leq 2 L^2 \gamma \mathbb{E}\| \bs{G}^t \|^2 + 2c\eta_l^2L^2K \sigma_l^2. \\
\end{aligned}
\end{equation}

Combining \eqref{eq:fedavg_descent} and \eqref{tmp:t_1}, we have
\begin{equation}\label{eq:fedavg_descent_2}
\begin{aligned}
\mathbb{E}\|\nabla f(\bs{x}^t)\|^2 & \leq \frac{2(f(\bs{x}^{t}) - \mathbb{E}\left[f(\bs{x}^{t+1})\right])}{\tilde{\eta}} + 2c\eta_l^2L^2K \sigma_l^2 +
(2L^2 \gamma + \frac{L \tilde{\eta}}{\eta_l^2 K^2}) \mathbb{E}[\|\bs{G}^t\|^2].
\end{aligned}
\end{equation}

Then, we investigate $\mathbb{E}\| \bs{G}^t \|^2$ using Assumption~\ref{asp:client_consunsus}. Straightforwardly, we know that
$$
\mathbb{E}\| \bs{G}^t \|^2 = \mathbb{V}[\bs{G}^t] + \| \mathbb{E}[\boldsymbol{G}^t]\|^2 = \mathbb{V}[\bs{G}^t] + \rho^t(\bs{\lambda}).
$$

Here, the variance of $\bs{G}^t$ is taken with respect to random noise in stochastic local updates. The upper bound only depends on the dynamics of local SGD, which has been well understood in literature~\citep{khaled2020tighter}. Concretely, author of \cite{khaled2020tighter} show that 
$$
\begin{aligned}
\mathbb{V}[\bs{G}^t] & = \mathbb{E}\| \bs{G}^t  - \mathbb{E}[\bs{G}^t]\|^2 = \mathbb{E}\|\sum_{i=1}^N \bs{\lambda}_i (\bs{g}_i^t - \mathbb{E}[\bs{g}_i^t])\|^2 \\
& \leq \sum_{i=1}^N \bs{\lambda}_i \mathbb{E}\|\bs{g}_i^t - \mathbb{E}[\bs{g}_i^t]\|^2 \leq K\eta_l^2 \sigma_l^2.
\end{aligned}
$$

Hence, we have the descent form as
\begin{equation}\label{eq:fedavg_sta_descent}
\begin{aligned}
\mathbb{E}\|\nabla f(\bs{x}^t)\|^2 & \leq \frac{2(f(\bs{x}^{t}) - \mathbb{E}\left[f(\bs{x}^{t+1})\right])}{\tilde{\eta}} + 2c\eta_l^2L^2K \sigma_l^2 +
(2L^2 \gamma + \frac{L \tilde{\eta}}{\eta_l^2 K^2} ) (K\eta_l^2 \sigma_l^2 + \rho^t(\bs{\lambda})) \\
& \leq \frac{2(f(\bs{x}^{0}) - \mathbb{E}\left[f(\bs{x}^{T})\right])}{T K \eta} + 2(c + \gamma)\eta_l^2L^2K\sigma_l^2 + \eta L\sigma_l^2 + 
\eta L (\frac{2 L \gamma}{\eta} + \frac{1}{\eta_l^2 K}) \rho^t(\bs{\lambda}) \\
& \leq \frac{2(f(\bs{x}^{0}) - \mathbb{E}\left[f(\bs{x}^{T})\right])}{T K \eta} + 2(c + \gamma)\eta_l^2L^2K\sigma_l^2 + \eta L\sigma_l^2 + 
\eta L (1+ \frac{\gamma}{4\eta_g}) \frac{\rho^t(\bs{\lambda})}{\eta_l^2 K} \\
\end{aligned}
\end{equation}
where we define $\eta = \eta_l\eta_g $ and use $\eta_l \leq \frac{1}{8LK}$ from Lemma~\ref{lemma:local_drift}.
Taking averaging of both sides from time $t=0$ to $T-1$, we obtain
\begin{equation}\label{eq:convergence_sta_fedavg}
\frac{1}{T}\sum_{t=0}^{T-1}\mathbb{E}\|\nabla f(\bs{x}^t)\|^2 \leq \frac{2(f(\bs{x}^{0}) - \mathbb{E}\left[f(\bs{x}^{T})\right])}{T K \eta} + 2(c + \gamma)\eta_l^2L^2K\sigma_l^2 + 
\eta L \sigma_l^2 + \eta L (1 + \frac{\gamma}{4\eta_g}) \frac{1}{T}\sum_{t=0}^{T-1} \frac{\rho^t(\bs{\lambda})}{\eta_l^2 K},
\end{equation}
which concludes the proof.

\subsection{Proof of Corollary~\ref{cor:adaptive_fedavg}}

Now, we investigate the convergence of FedAvg when the global objective weight $\bs{\lambda}$ and the applied aggregation weight $\tilde{\bs{\lambda}}$ are not identical. The update rule on the server side becomes
$$
\bs{x}^{t+1} = \bs{x}^{t} -\tilde{\eta}\tilde{\bs{G}}^t, \quad \tilde{\bs{G}}^t = \sum_{i=1}^N \tilde{\bs{\lambda}}_i^t \bs{g}_i^t,
$$
where $\tilde{\bs{\lambda}}^t$ is given by any aggregation strategy on FL server. Noting that original weights $\bs{\lambda}$ is used to define global objective, inducing $\nabla f(\bs{x})= \sum_{i=}^N \bs{\lambda}_i \nabla f_i(\bs{x})$.

For ease of writing, we investigate the normalized update rule:
$$
\bs{x}^{t+1} = \bs{x}^{t} -\tilde{\eta}\bar{\bs{G}}^t, \quad \bar{\bs{G}}^t = \frac{1}{\eta_l K} \tilde{\bs{G}}^t.
$$
Using the smoothness, we have:
$$
\begin{aligned}
f\left(\bs{x}^{t+1}\right) & \leq f(\bs{x}^t) - \frac{\tilde{\eta}}{2} \|\nabla f(\bs{x}^t)\|^2 + \frac{\tilde{\eta}}{2} \|\nabla f(\bs{x}^t) - \bar{\bs{G}}^t\|^2 + \frac{L}{2}\tilde{\eta}^2\left\|\bar{\bs{G}}^t\right\|^2 \\ 
\end{aligned}
$$

Taking full expectation over randomness at time step $t$ on both sides, we have:
\begin{equation}\label{eq:fedavg_adp_descent}
\begin{aligned}
\mathbb{E}\left[f(\bs{x}^{t+1})\right] - f(\bs{x}^{t}) & \leq - \frac{\tilde{\eta}}{2} \mathbb{E}\|\nabla f(\bs{x}^t)\|^2 + \frac{\tilde{\eta}}{2}\mathbb{E}\|\nabla f(\bs{x}^t) - \bar{\bs{G}}^t\|^2 + \frac{L}{2} \tilde{\eta}^2 \mathbb{E}[\|\bar{\bs{G}}^t\|^2] \\
& \leq - \frac{\tilde{\eta}}{2} \mathbb{E}\|\nabla f(\bs{x}^t)\|^2 + \frac{\tilde{\eta}}{2}\mathbb{E}\|\nabla f(\bs{x}^t) \pm \hat{\bs{G}}^t - \bar{\bs{G}}^t\|^2 + \frac{L}{2} \tilde{\eta}^2 \mathbb{E}[\|\bar{\bs{G}}^t\|^2] \\
& \leq - \frac{\tilde{\eta}}{2} \mathbb{E}\|\nabla f(\bs{x}^t)\|^2 + \tilde{\eta}\mathbb{E}\|\nabla f(\bs{x}^t) - \hat{\bs{G}}^t\|^2 + \tilde{\eta} \mathbb{E}\|\hat{\bs{G}}^t - \bar{\bs{G}}^t\|^2 + \frac{L}{2} \tilde{\eta}^2 \mathbb{E}[\|\bar{\bs{G}}^t\|^2] 
\end{aligned}
\end{equation}

To compare with vanilla FedAvg (static weight), we investigate the gap between expected pseudo-gradient $\bs{G}^t$ and applied pseudo-gradient $\tilde{\bs{G}}^t$. By definition, we know
$$
\bs{G}^t - \tilde{\bs{G}}^t = \sum_{i=1}^N(\bs{\lambda}_i - \tilde{\bs{\lambda}}_i) \bs{g}_i^t = \sum_{i=1}^N\frac{(\bs{\lambda}_i - \tilde{\bs{\lambda}}_i)}{\sqrt{\bs{\lambda}_i}} \sqrt{\bs{\lambda}_i} \bs{g}_i^t.
$$

Then, applying Cauchy-Schwarz inequality, it induces that
\begin{equation}\label{eq:bouned_weights_gap}
\begin{aligned}
\mathbb{E}\|\hat{\bs{G}}^t - \bar{\bs{G}}^t\|^2 = \frac{1}{\eta_l^2 K^2} \mathbb{E}\|\bs{G}^t - \tilde{\bs{G}}^t\|^2 \leq \frac{1}{\eta_l^2 K^2}\left[\sum_{i=1}^N \frac{(\bs{\lambda}_i - \tilde{\bs{\lambda}}_i)^2}{\bs{\lambda}_i}\right] \cdot \left[ \sum_{i=1}^N \bs{\lambda}_i \left\|\bs{g}_i^t\right\|^2\right] = \frac{\chi_{\bs{\lambda}\|\tilde{\bs{\lambda}}}^2}{\eta_l^2 K^2} \sum_{i=1}^N \bs{\lambda}_i \left\|\bs{g}_i^t\right\|^2,
\end{aligned}
\end{equation}
where we define $\chi_{\bs{\lambda}\|\tilde{\bs{\lambda}}}^2 = \sum_{i=1}^N \frac{(\bs{\lambda}_i - \tilde{\bs{\lambda}}_i)^2}{\bs{\lambda}_i}$. Then, reorganizing the terms in \eqref{eq:fedavg_adp_descent}, we have
$$
\begin{aligned}
\mathbb{E}\|\nabla f(\bs{x}^t)\|^2 & \leq \frac{2(f(\bs{x}^{t}) - \mathbb{E}\left[f(\bs{x}^{t+1}))\right]}{K \eta} + 2\mathbb{E}\|\nabla f(\bs{x}^t) - \hat{\bs{G}}^t\|^2 + 2 \mathbb{E}\|\hat{\bs{G}}^t - \bar{\bs{G}}^t\|^2 + L \tilde{\eta} \mathbb{E}[\|\bar{\bs{G}}^t\|^2] \\
& \leq \frac{2(f(\bs{x}^{t}) - \mathbb{E}\left[f(\bs{x}^{t+1}))\right]}{K \eta} +   (4L^2+\frac{2\chi_{\bs{\lambda}\|\tilde{\bs{\lambda}}}^2}{\eta_l^2 K^2}) \sum_{i=1}^N \bs{\lambda}_i\mathbb{E}\left\| \bs{g}_i^t\right\|^2 + 4c\eta_l^2L^2K\sigma_l^2  + L \tilde{\eta} \mathbb{E}[\|\bar{\bs{G}}^t\|^2], \\
\end{aligned}
$$
where the last inequality uses \eqref{tmp:t_1} and \eqref{eq:bouned_weights_gap}.

Then, we use Assumption~\ref{asp:norm_relation} to have
$$
\begin{aligned}
\mathbb{E}\|\nabla f(\bs{x}^t)\|^2 & \leq \frac{2(f(\bs{x}^{t}) - \mathbb{E}\left[f(\bs{x}^{t+1}))\right]}{K \eta} +  (4L^2+\frac{2\chi_{\bs{\lambda}\|\tilde{\bs{\lambda}}}^2}{\eta_l^2 K^2}) \gamma \mathbb{E}\| \tilde{\bs{G}}^t \|^2 + 4c\eta_l^2L^2K\sigma_l^2 + L \tilde{\eta} \mathbb{E}[\|\bar{\bs{G}}^t\|^2] \\
& \leq \frac{2(f(\bs{x}^{t}) - \mathbb{E}\left[f(\bs{x}^{t+1}))\right]}{K \eta} +  (4L^2+\frac{2\chi_{\bs{\lambda}\|\tilde{\bs{\lambda}}}^2}{\eta_l^2 K^2}) \gamma \mathbb{E}\| \tilde{\bs{G}}^t \|^2 + 4c\eta_l^2L^2K\sigma_l^2 + L \tilde{\eta} \mathbb{E}[\|\bar{\bs{G}}^t\|^2] \\
& \leq \frac{2(f(\bs{x}^{t}) - \mathbb{E}\left[f(\bs{x}^{t+1}))\right]}{K \eta} + 4c\eta_l^2L^2K\sigma_l^2 +
((4L^2 + \frac{2\chi_{\bs{\lambda}\|\tilde{\bs{\lambda}}}^2}{\eta_l^2 K^2})\gamma + \frac{L \tilde{\eta}}{\eta_l^2 K^2}) \mathbb{E}[\|\tilde{\bs{G}}^t\|^2] \\
& \leq \frac{2(f(\bs{x}^{t}) - \mathbb{E}\left[f(\bs{x}^{t+1}))\right]}{K \eta} + 4c\eta_l^2L^2K\sigma_l^2 + ((4L^2 + \frac{2\chi_{\bs{\lambda}\|\tilde{\bs{\lambda}}}^2}{\eta_l^2 K^2})\gamma + \frac{L \tilde{\eta}}{\eta_l^2 K^2}) \cdot (K\eta_l^2 \sigma_l^2 + \rho^t(\tilde{\bs{\lambda}})).
\end{aligned}
$$

Now, we conducting similar operations in \eqref{eq:fedavg_sta_descent} to obtain
$$
\mathbb{E}\|\nabla f(\bs{x}^t)\|^2 \leq \frac{2(f(\bs{x}^{t}) - \mathbb{E}\left[f(\bs{x}^{t+1}))\right]}{K \eta} + 4((1+\frac{\chi_{\bs{\lambda}\|\tilde{\bs{\lambda}}}^2}{2\eta_l^2L^2K^2})\gamma + c)\eta_l^2L^2K \sigma_l^2 + \eta L \sigma_l^2 + \eta L ((\frac{L}{2\eta_g} + \frac{2\chi_{\bs{\lambda}\|\tilde{\bs{\lambda}}}^2}{\eta KL})\gamma + 1) \frac{\rho^t(\tilde{\bs{\lambda}})}{\eta_l^2 K}.
$$
Taking averaging of both sides from time $t=0$ to $T-1$ and defining $\tilde{\chi} = \frac{1}{T}\sum_{t=0}^{T-1} \chi_{\bs{\lambda}\|\tilde{\bs{\lambda}}}^2$ for notation simplicity, we obtain
\begin{equation}\label{eq:convergence_adap_fedavg}
\begin{aligned}
\frac{1}{T}\sum_{t=0}^{T-1} \mathbb{E}\|\nabla f(\bs{x}^t)\|^2 & \leq \frac{2(f(\bs{x}^{0}) - \mathbb{E}\left[f(\bs{x}^{T})\right])}{T K \eta} + \eta L \sigma_l^2 + 4(c + \gamma +\frac{\tilde{\chi}\gamma}{2\eta_l^2L^2K^2}) \cdot \eta_l^2L^2K \sigma_l^2 \\
&\quad  + \eta L (1 + \frac{L\gamma}{2\eta_g} + \frac{2\tilde{\chi}\gamma}{
\eta KL}) \frac{1}{T}\sum_{t=0}^{T-1} \frac{\rho^t(\tilde{\bs{\lambda}})}{\eta_l^2 K}.
\end{aligned}
\end{equation}

In the main paper, we mainly discuss the differences between the last term in Eq.\eqref{eq:convergence_adap_fedavg} and Eq.\eqref{eq:convergence_sta_fedavg}, emphasizing the benefits of adaptive global aggregation.

\subsection{Convergence Analysis of FedAWARE with Partial Participation}

We recall the update rule of Algorithm~\ref{alg:gdm} is:
$$
\bs{x}^{t+1} = \bs{x}^{t} - \eta \bs{d}^t,
$$
where 
$$
\bs{d}^t = \sum_{i=1}^N \tilde{\lambda}^t_i \bs{m}_i^t, \text{s.t.} \; \tilde{\lambda}^t = \underset{\lambda}{\arg \min }\left\|\sum_{i=1}^N \bs{\lambda}_i \bs{m}_i^t\right\|^2 .
$$

We rewrite the update rule with normalized pseudo-gradient:
$$
\bs{x}^{t+1} = \bs{x}^{t} - \tilde{\eta} \hat{\bs{d}}^t,
$$
where $\tilde{\eta} = K \eta_l\eta_g$ and $\hat{\bs{d}}^t = \bs{d}^t / \eta_l K$. 
Using the smoothness, we have:
$$
\begin{aligned}
f\left(\bs{x}^{t+1}\right) & \leq f(\bs{x}^t) - \frac{\tilde{\eta}}{2} \|\nabla f(\bs{x}^t)\|^2 + \frac{\tilde{\eta}}{2} \|\nabla f(\bs{x}^t) - \hat{\bs{d}}^t \|^2 + \frac{L}{2}\tilde{\eta}^2\left\|\hat{\bs{d}}^t\right\|^2. \\ 
\end{aligned}
$$

Taking full expectation over randomness at time step $t$ on both sides, we have:
\begin{equation}\label{eq:original}
\begin{aligned}
\mathbb{E}\left[f(\bs{x}^{t+1})\right] - f(\bs{x}^{t}) & \leq - \frac{\tilde{\eta}}{2} \mathbb{E} \|\nabla f(\bs{x}^t)\|^2 + \frac{\tilde{\eta}}{2}\underbrace{\mathbb{E}\|\nabla f(\bs{x}^t) - \hat{\bs{d}}^t \|^2}_{T_1} + \frac{L}{2} \tilde{\eta}^2 \underbrace{\mathbb{E}[\|\hat{\bs{d}}^t\|^2]}_{T_2}.
\end{aligned}
\end{equation}

\textbf{Bounding $T_1$  }
Following the definition, we have
$$
\begin{aligned}
\mathbb{E}\|\nabla f(\bs{x}^t) - \hat{\bs{d}}^t \|^2 & \leq \mathbb{E}\|\nabla f(\bs{x}^t) \pm  \hat{\bs{G}}^t - \hat{\bs{d}}^t \|^2 \\
& \leq 2\mathbb{E}\|\nabla f(\bs{x}^t) -  \hat{\bs{G}}^t\|^2 +   2\mathbb{E}\|\hat{\bs{G}}^t - \hat{\bs{d}}^t \|^2 \\
& \leq 2\mathbb{E}\|\nabla f(\bs{x}^t) -  \hat{\bs{G}}^t\|^2 +   2\frac{1}{\eta_l^2 K^2}\mathbb{E}\|\sum_{i=1}^N \bs{\lambda}_i \bs{g}_i^t - \sum_{i=1}^N \tilde{\bs{\lambda}}_i \bs{m}_i^t \|^2 \\
& \leq 2\mathbb{E}\|\nabla f(\bs{x}^t) -  \hat{\bs{G}}^t\|^2 +   2\frac{1}{\eta_l^2 K^2}\mathbb{E}\|\sum_{i=1}^N \bs{\lambda}_i \bs{g}_i^t \pm \sum_{i=1}^N \tilde{\bs{\lambda}}_i \bs{g}_i^t -  \sum_{i=1}^N \tilde{\bs{\lambda}}_i \bs{m}_i^t \|^2 \\
& \leq 2\underbrace{\mathbb{E}\|\nabla f(\bs{x}^t) -  \hat{\bs{G}}^t\|^2}_{\text{Eq.\eqref{tmp:t_1}}} +  4\underbrace{\frac{1}{\eta_l^2 K^2}\mathbb{E}\|\sum_{i=1}^N \bs{\lambda}_i \bs{g}_i^t - \sum_{i=1}^N \tilde{\bs{\lambda}}_i \bs{g}_i^t\|^2}_{\text{Eq.\eqref{eq:bouned_weights_gap}}} + 4\frac{1}{\eta_l^2 K^2}\mathbb{E}\|\sum_{i=1}^N \tilde{\bs{\lambda}}_i \bs{g}_i^t -  \sum_{i=1}^N \tilde{\bs{\lambda}}_i \bs{m}_i^t \|^2, \\
\end{aligned}
$$
where the first and second terms have been bounded. 

We know
\begin{equation}\label{eq:bouned_momentum_gap}
\begin{aligned}
\mathbb{E}\|\sum_{i=1}^N \tilde{\bs{\lambda}}_i \bs{g}_i^t - \sum_{i=1}^N \tilde{\bs{\lambda}}_i \bs{m}_i^t \|^2 & = \mathbb{E}\|\sum_{i=1}^N \frac{\tilde{\bs{\lambda}}_i}{\sqrt{\bs{\lambda}_i}}\cdot \sqrt{\bs{\lambda}_i}(\bs{g}_i^t - \bs{m}_i^t) \|^2 \leq \sum_{i=1}^N\frac{\tilde{\bs{\lambda}}_i^2}{\bs{\lambda}_i} \cdot \sum_{i=1}^N \bs{\lambda}_i \mathbb{E}\|\bs{g}_i^t - \bs{m}_i^t\|^2 \\
\end{aligned}
\end{equation}
by Cauchy-Schwarz inequality. Now, we turn to bound the approximation error of moving-averaged local updates. Letting $p_i = \text{Prob}(i \in S^t)$ be the probability of $i$-th clients be selected at the round, we have 
$$
\begin{aligned}
\mathbb{E}\|\bs{m}_i^t - \bs{g}_i^t\|^2 & = \mathbb{E}\|(1-p_i)\bs{m}_i^{t-1}+p_i ((1-\alpha) \bs{m}_i^{t-1} + \alpha \bs{g}_i^t) - \bs{g}_i^t\|^2 \\
& = \mathbb{E}\| (1-\alpha p_i) \bs{m}_i^{t-1} \pm (1-\alpha p_i)\bs{g}_i^{t-1} - (1-\alpha p_i) \bs{g}_i^t\|^2 \\
& \leq (1-\alpha p_i)^2 \mathbb{E}\|\bs{m}_i^{t-1}-\bs{g}_i^{t-1}\|^2 + (1-\alpha p_i)^2 \mathbb{E}\| \bs{g}_i^{t-1}- \bs{g}_i^t\|^2 \\
& \leq (1-\alpha p_i)^2 \mathbb{E}\|\bs{m}_i^{t-1}-\bs{g}_i^{t-1}\|^2 + (1-\alpha p_i)^2 \mathbb{E}\| \bs{g}_i^{t-1}\|^2 + (1-\alpha p_i)^2 \mathbb{E}\|\bs{g}_i^t\|^2.
\end{aligned}
$$

Then, letting $\beta_i = (1-\alpha p_i)^2 \ll 1$ for simple notion and unrolling the recursion from time $0$ to $t$ with the initialization such that $\mathbb{E}\|\bs{m}_i^0 - \bs{g}_i^0\|^2=0$, we have 
$$
\mathbb{E}\|\bs{m}_i^t - \bs{g}_i^t\|^2 \leq \sum_{\tau=0}^{t-1} \beta_i^{t-\tau} (\mathbb{E}\| \bs{g}_i^{\tau}\|^2 + \mathbb{E}\|\bs{g}_i^{\tau+1}\|^2).
$$
Noting that $\beta \ll 1$, we can exploit this small parameter to simplify the expression by observing that the terms involving $\beta^{t-\tau}$ decay rapidly. This allows us to approximate the sum by focusing primarily on the latest terms $t$, effectively reducing the accumulation effect. Therefore, the accumulation can be neglected, and the bound simplifies to
$$
\mathbb{E}\|\bs{m}_i^t - \bs{g}_i^t\|^2 \leq 2 \beta (\mathbb{E}\| \bs{g}_i^{t-1}\|^2 + \mathbb{E}\|\bs{g}_i^{t}\|^2) \leq 2 \beta \max(\mathbb{E}\| \bs{g}_i^{t-1}\|^2, \mathbb{E}\|\bs{g}_i^{t}\|^2) \leq 2 \beta \mathbb{E}\|\bs{g}_i^{t}\|^2,
$$ 
without loss of generality. Combining the above terms, we obtain
$$
\begin{aligned}
T_1 & \leq 4 L^2 \gamma \mathbb{E}\| \bs{G}^t \|^2 + 4c\eta_l^2L^2K \sigma_l^2 + 4 \frac{\chi_{\bs{\lambda}\|\tilde{\bs{\lambda}}}^2}{\eta_l^2 K^2} \sum_{i=1}^N \bs{\lambda}_i \left\|\bs{g}_i^t\right\|^2 + 8 \frac{\beta}{\eta_l^2 K^2} \sum_{i=1}^N\frac{\tilde{\bs{\lambda}}_i^2}{\bs{\lambda}_i} \cdot \sum_{i=1}^N \bs{\lambda}_i \mathbb{E}\|\bs{g}_i^{t}\|^2 \\
& = 4\left(L^2 + \frac{\chi_{\bs{\lambda}\|\tilde{\bs{\lambda}}}^2}{\eta_l^2 K^2} + \frac{2\beta}{\eta_l^2 K^2} \sum_{i=1}^N\frac{\tilde{\bs{\lambda}}_i^2}{\bs{\lambda}_i}\right) \gamma \mathbb{E}\|\tilde{\bs{G}}^t\|^2 + 4c\eta_l^2L^2K \sigma_l^2 \\
& \leq 4\left(L^2 + \frac{(1+2\beta)\chi_{\bs{\lambda}\|\tilde{\bs{\lambda}}}^2}{\eta_l^2 K^2}\right) \gamma \mathbb{E}\|\tilde{\bs{G}}^t\|^2 + 4c\eta_l^2L^2K \sigma_l^2 \\
\end{aligned}
$$
where the last inequality assumes $\sum_{i=1}^N\frac{\tilde{\bs{\lambda}}_i^2}{\bs{\lambda}_i} \leq \chi_{\bs{\lambda}\|\tilde{\bs{\lambda}}}^2 = \sum_{i=1}^N \frac{(\bs{\lambda}_i - \tilde{\bs{\lambda}}_i)^2}{\bs{\lambda}_i} $ without loss of generality.

\textbf{Bounding $T_2$  }
Following the definition, we have
$$
\begin{aligned}
\mathbb{E}\|\hat{\bs{d}}^t\|^2 & = \frac{1}{\eta_l^2K^2}\mathbb{E}\|\sum_{i=1}^N \tilde{\bs{\lambda}}_i \bs{m}_i^t \pm \sum_{i=1}^N \tilde{\bs{\lambda}}_i \bs{g}_i^t\|^2 \\
& \leq \frac{2}{\eta_l^2K^2}\mathbb{E}\|\sum_{i=1}^N \tilde{\bs{\lambda}}_i (\bs{m}_i^t - \bs{g}_i^t)\|^2 + \frac{2}{\eta_l^2K^2}\mathbb{E}\|\sum_{i=1}^N \tilde{\bs{\lambda}}_i \bs{g}_i^t\|^2 \\
& = \frac{2}{\eta_l^2K^2}\underbrace{\mathbb{E}\|\sum_{i=1}^N \tilde{\bs{\lambda}}_i (\bs{m}_i^t - \bs{g}_i^t)\|^2}_{\text{Eq.\eqref{eq:bouned_momentum_gap}}} + \frac{2}{\eta_l^2K^2}\mathbb{E}\|\tilde{\bs{G}}^t\|^2 \\
\end{aligned}
$$

Substituting corresponding terms, we have
$$
\begin{aligned}
T_2 & \leq 8 \beta \sum_{i=1}^N\frac{\tilde{\bs{\lambda}}_i^2}{\bs{\lambda}_i} \cdot \sum_{i=1}^N \bs{\lambda}_i \mathbb{E}\|\bs{g}_i^{t}\|^2  + 2\mathbb{E}\|\tilde{\bs{G}}^t\|^2 \\
& \leq \frac{8}{\eta_l^2K^2} \beta \chi_{\bs{\lambda}\|\tilde{\bs{\lambda}}}^2 \gamma \mathbb{E}\|\tilde{\bs{G}}^t\|^2  + \frac{2}{\eta_l^2K^2}\mathbb{E}\|\tilde{\bs{G}}^t\|^2 \\
& = 2 \frac{1+4\beta \chi_{\bs{\lambda}\|\tilde{\bs{\lambda}}}^2 \gamma}{\eta_l^2K^2}\mathbb{E}\|\tilde{\bs{G}}^t\|^2
\end{aligned}
$$

\textbf{Putting together  } Substituting upper bound of $T_1$ and $T_2$ in \eqref{eq:original}, we have

$$
\begin{aligned}
\mathbb{E} \|\nabla f(\bs{x}^t)\|^2  & \leq \frac{2(f(\bs{x}^{t}) - \mathbb{E}\left[f(\bs{x}^{t+1}))\right]}{\eta K} + 4c\eta_l^2L^2K \sigma_l^2 \\
& \quad + 4\left(L^2 + \frac{(1+2\beta)\chi_{\bs{\lambda}\|\tilde{\bs{\lambda}}}^2}{\eta_l^2 K^2}\right) \gamma \mathbb{E}\|\tilde{\bs{G}}^t\|^2 + L\tilde{\eta} \cdot 2 \frac{1+4\beta \chi_{\bs{\lambda}\|\tilde{\bs{\lambda}}}^2 \gamma}{\eta_l^2K^2}\mathbb{E}\|\tilde{\bs{G}}^t\|^2 \\
& \leq \frac{2(f(\bs{x}^{t}) - \mathbb{E}\left[f(\bs{x}^{t+1}))\right]}{\eta K} + 4c\eta_l^2L^2K \sigma_l^2 \\
& \quad + 4\left(L^2 + \frac{(1+2\beta)\chi_{\bs{\lambda}\|\tilde{\bs{\lambda}}}^2}{\eta_l^2 K^2}\right) \gamma \cdot (K\eta_l^2 \sigma_l^2 + \rho^t(\tilde{\bs{\lambda}})) + L\tilde{\eta} \cdot 2 \frac{1+4\beta \chi_{\bs{\lambda}\|\tilde{\bs{\lambda}}}^2 \gamma}{\eta_l^2K^2} \cdot (K\eta_l^2 \sigma_l^2 + \rho^t(\tilde{\bs{\lambda}})) \\
& \leq \frac{2(f(\bs{x}^{t}) - \mathbb{E}\left[f(\bs{x}^{t+1}))\right]}{\eta K} + 4(c+\left(1 + \frac{(1+2\beta)\chi_{\bs{\lambda}\|\tilde{\bs{\lambda}}}^2}{\eta_l^2 L^2K^2}\right) \gamma) \cdot \eta_l^2L^2K \sigma_l^2 + 2L(1+4\beta \chi_{\bs{\lambda}\|\tilde{\bs{\lambda}}}^2 \gamma) \cdot \eta\sigma_l^2   \\
& \quad + 2 \eta L \cdot \left(1 + \frac{L\gamma}{8\eta_g} +  \frac{2(1+2\beta)\chi_{\bs{\lambda}\|\tilde{\bs{\lambda}}}^2 \gamma}{\eta KL} + 4\beta \chi_{\bs{\lambda}\|\tilde{\bs{\lambda}}}^2 \gamma \right) \cdot  \frac{\rho^t(\tilde{\bs{\lambda}})}{\eta_l^2 K}.
\end{aligned}
$$

Taking averaging of both sides from time $t=0$ to $T-1$, and defining $\tilde{\chi} = \frac{1}{T}\sum_{t=0}^{T-1} \chi_{\bs{\lambda}\|\tilde{\bs{\lambda}}}^2$ for notation simplicity, we obtain
\begin{equation}
\begin{aligned}
\frac{1}{T}\sum_{t=0}^{T-1} \mathbb{E} \|\nabla f(\bs{x}^t)\|^2  & \leq \frac{2(f(\bs{x}^{0}) - \mathbb{E}\left[f(\bs{x}^{T}))\right]}{\eta T K} + 4\left(c+ \gamma + \frac{(1+2\beta)\tilde{\chi}\gamma}{\eta_l^2L^2K^2}\right) \cdot \eta_l^2L^2K \sigma_l^2 + 2L(1+4\beta\tilde{\chi} \gamma) \cdot \eta\sigma_l^2   \\
& \quad + 2 \eta L \cdot \left(1 + \frac{L\gamma}{8\eta_g} +  \frac{2(1+2\beta)\tilde{\chi} \gamma}{\eta KL} + 4\beta\tilde{\chi} \gamma \right) \frac{1}{T}\sum_{t=0}^{T-1} \frac{\rho^t(\tilde{\bs{\lambda}})}{\eta_l^2 K}.
\end{aligned}
\end{equation}

Compared with \eqref{eq:convergence_adap_fedavg}, we see that FedAWARE with partial participation only induces additional minor coefficients multiplied by $\beta$. As $\beta \ll 1$ in practice, we argue that FedAWARE is robust to partial client participation in theory. Moreover, empirical evidence in the main paper also proves this point.

\section{Experiment Details}

\subsection{Experiment Details}\label{app:exp_details}

\textbf{Platform  } The experiment implementations are supported by FedLab~\citep{JMLR:v24:22-0440}. Our experiments run on a Linux server with 4*2080Ti GPU.

\begin{figure}[h]
\centering
\subfigure[CIFAR-10]{\includegraphics[width=0.9\linewidth]{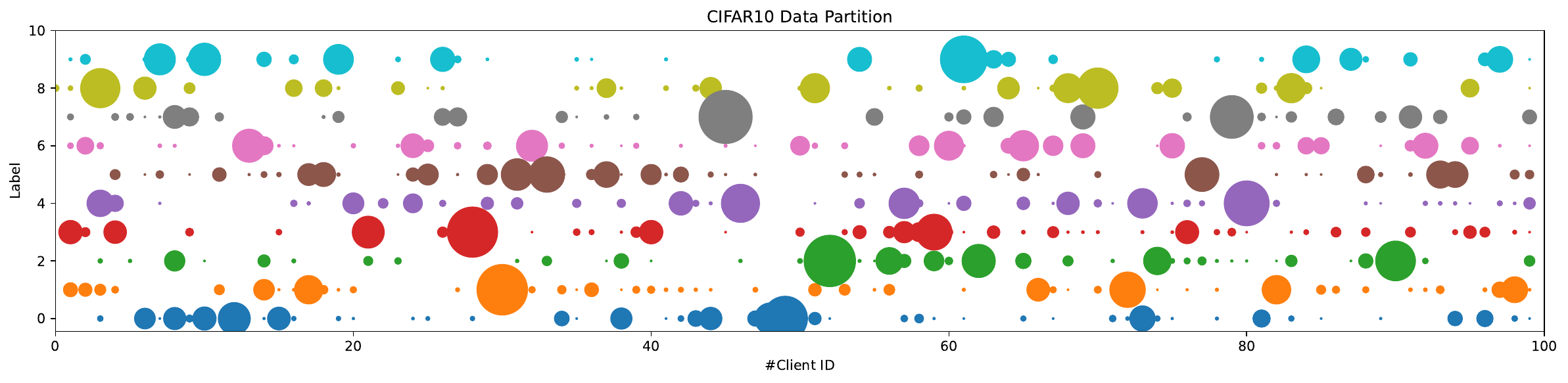}}
\subfigure[CIFAR-100]{\includegraphics[width=0.9\linewidth]{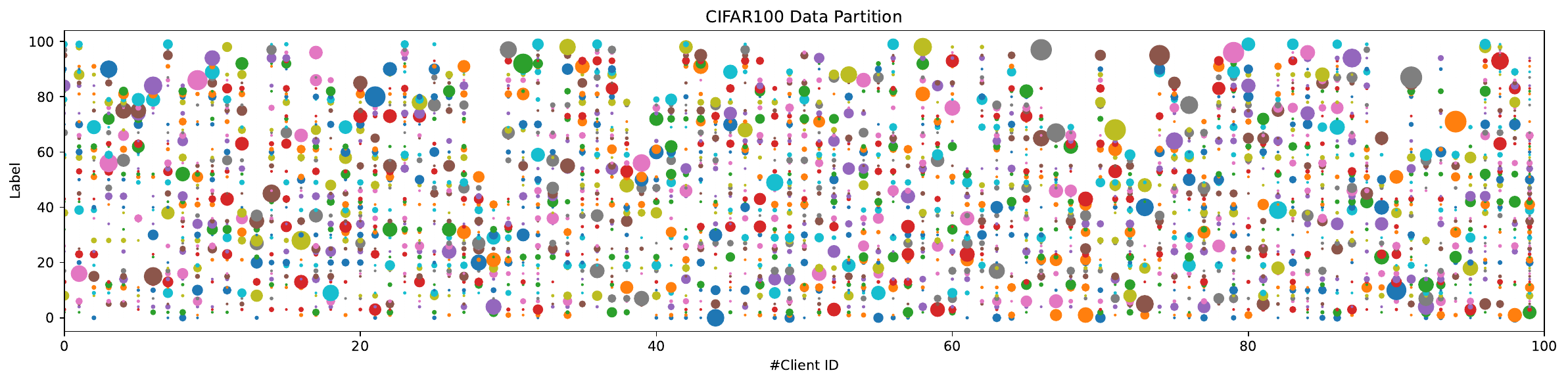}}
\subfigure[AGNews]{\includegraphics[width=0.9\linewidth]{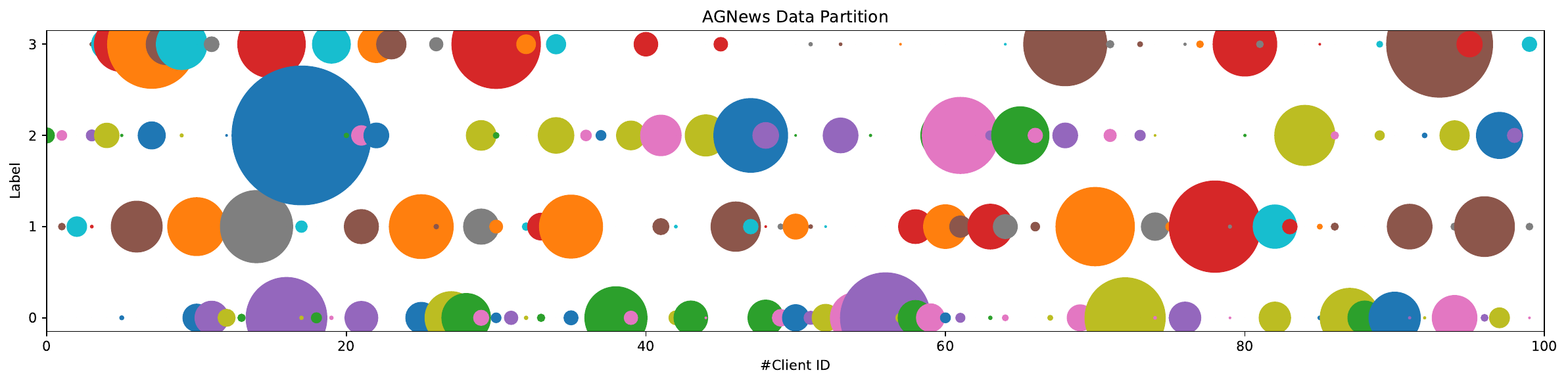}}
\caption{Visualization of data distribution.}\label{fig:data}
\end{figure}

\textbf{Data partition  } We present the data distribution of datasets in Figure~\ref{fig:data}.

\textbf{Algorithm-specified hyperparameters  }  We set the rate of client participation to be $10\%$, and use $\eta_g=1$ for FedAvg, FedAvgM, FedProx, FedCM, and FedAWARE. 
For the momentum parameter, FedAvgM is chosen from $\{0.7, 0.9, 0.97, 0.997\}$ following the original paper, and FedCM is chosen from $\{0.1, 0.3, 0.5, 0.7, 0.9\}$.
For weights of the penalty term in FedProx, we tune it from grid $\{0.01, 0.1, 1, 10\}$. For FedYogi, we set momentum parameter $\beta_1=0.9$, second-momentum parameter $\beta_2 = 0.99$, and adaptivity $\tau=10^{-4}$ following the original paper. Besides, We select $\eta_g$ for FedYogi and FedAMS by grid-searching tuning from $\{10^{-3}, 10^{-2.5}, 10^{-2}, \dots, 10^1\}$. 
The parameter of FedDyn is chosen among $\{0.1, 0.01, 0.001\}$ from the original paper. For FedAMS, we set $\beta_1=0.9, \beta_2 = 0.99$ follows the original paper. Then, we grid search for the best global learning rate $\eta_g = \{1, 10^{-1}, 10^{-2}, 10^{-3}, 10^{-4}\}$ and the best stabilization term $\epsilon = \{10^{-8}, 10^{-4}, 10^{-3}, 10^{-2}, 10^{-1}\}$. We set $\alpha=0.5$ for FedAWARE. 

\subsection{Additional Experimental Results}\label{app:exp_comparison}

\begin{figure*}[h]
\centering
\subfigure[CIFAR-10]{\includegraphics[width=0.465\linewidth]{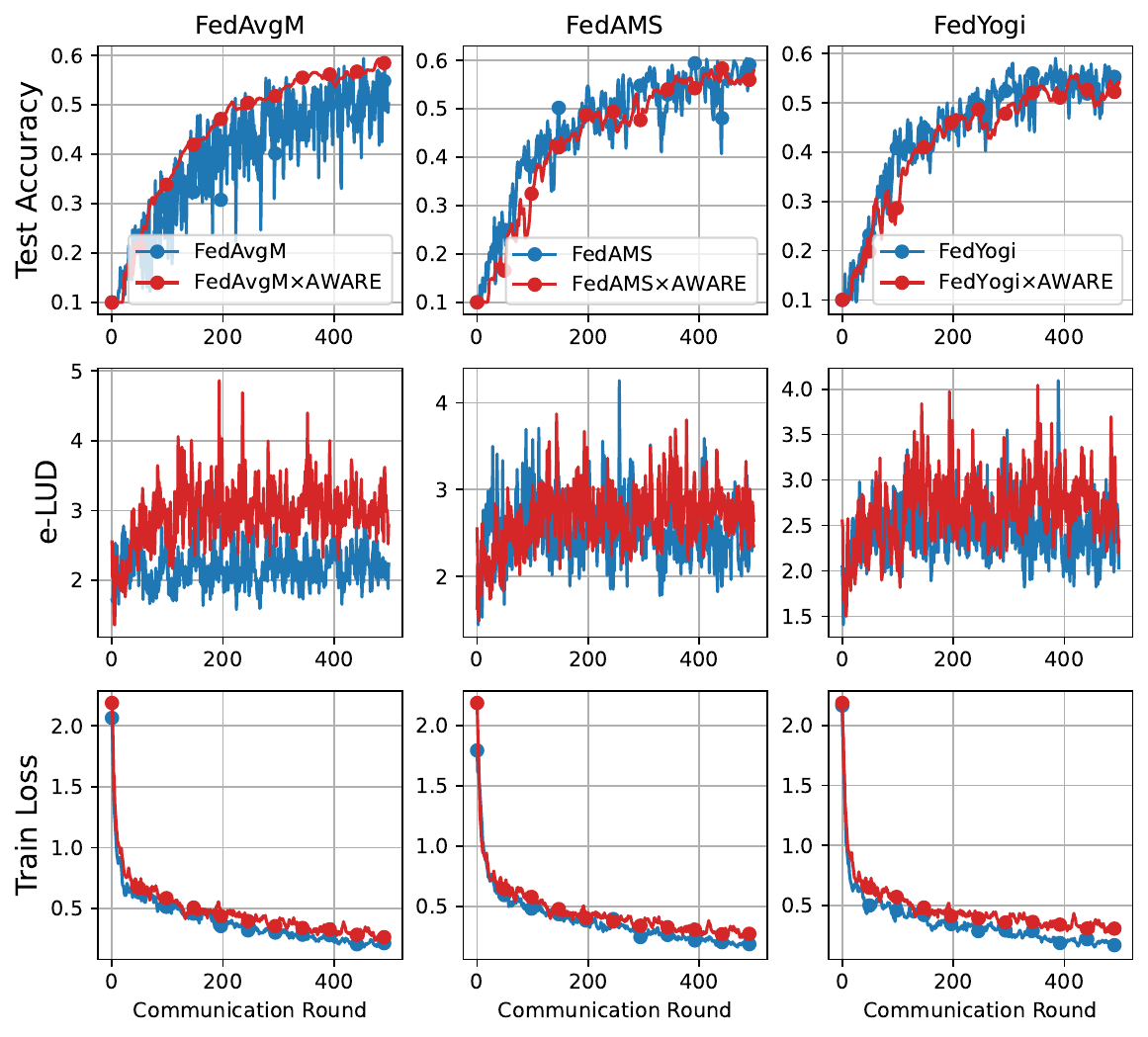}}
\subfigure[AGNews]{\includegraphics[width=0.465\linewidth]{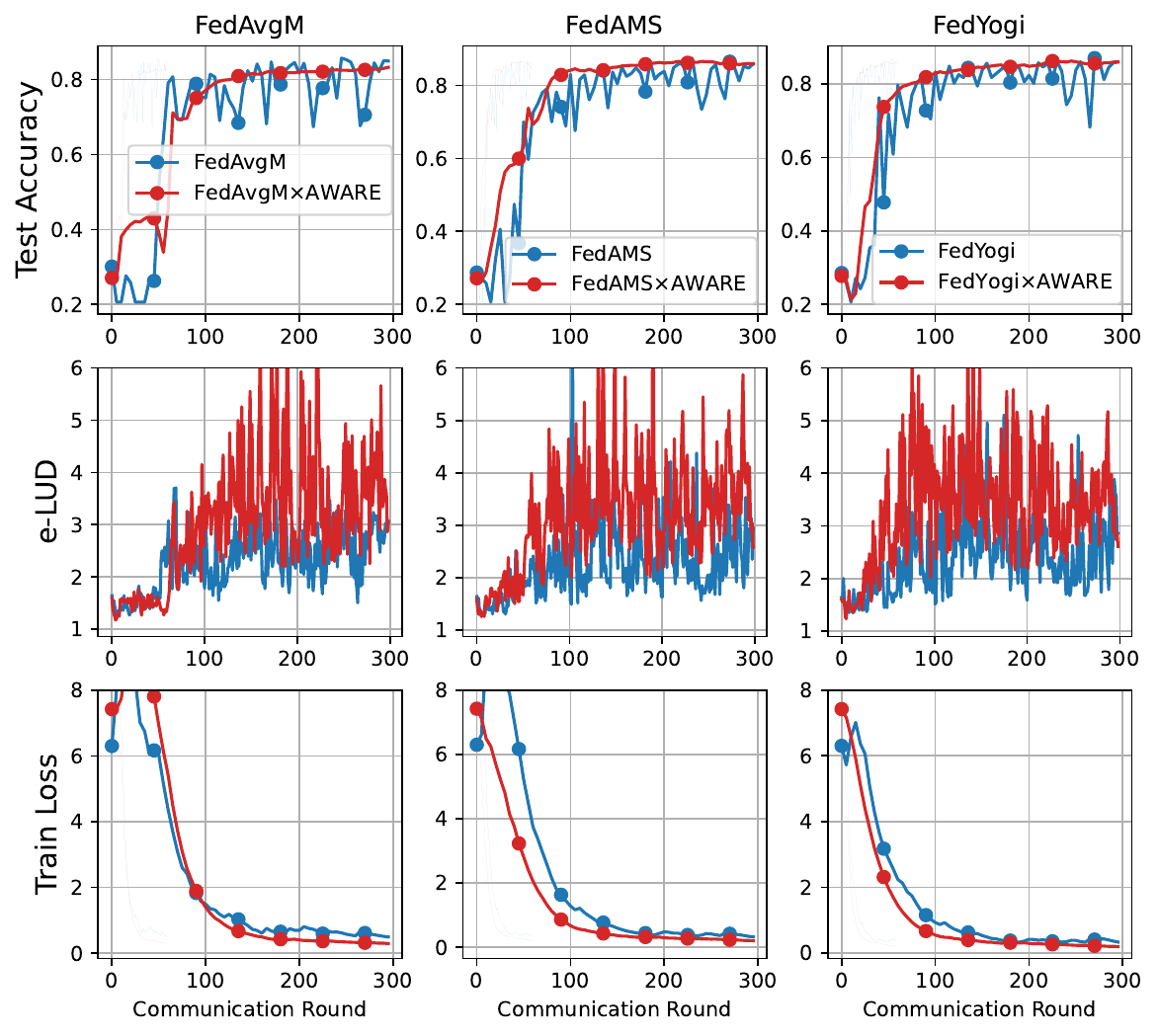}}
\caption{Training dynamics of FedAWARE extension.}\label{fig:extension}
\end{figure*}

\textbf{Experiments results on CIFAR-10 and AGNews  } 
We present the FedAWARE extension results of CIFAR-10 and AGNews in Figure~\ref{fig:extension}. We observe that raw federated algorithms test accuracy curves with heavy spike problems, indicating the heterogeneous data impacts on FL generalization. On both tasks, FedAWARE enlarges the e-LUD values of all algorithms and thus stabilizes the test accuracy curves. Moreover, connecting the results of CIFAR-100 in the main paper, we observe that the FedAWARE extension may reduce convergence slightly on CIFAR-10 and CIFAR-100 tasks while the convergence of AGNews tasks is improved.

\begin{figure}[h]
\centering
\includegraphics[width=0.8\linewidth]{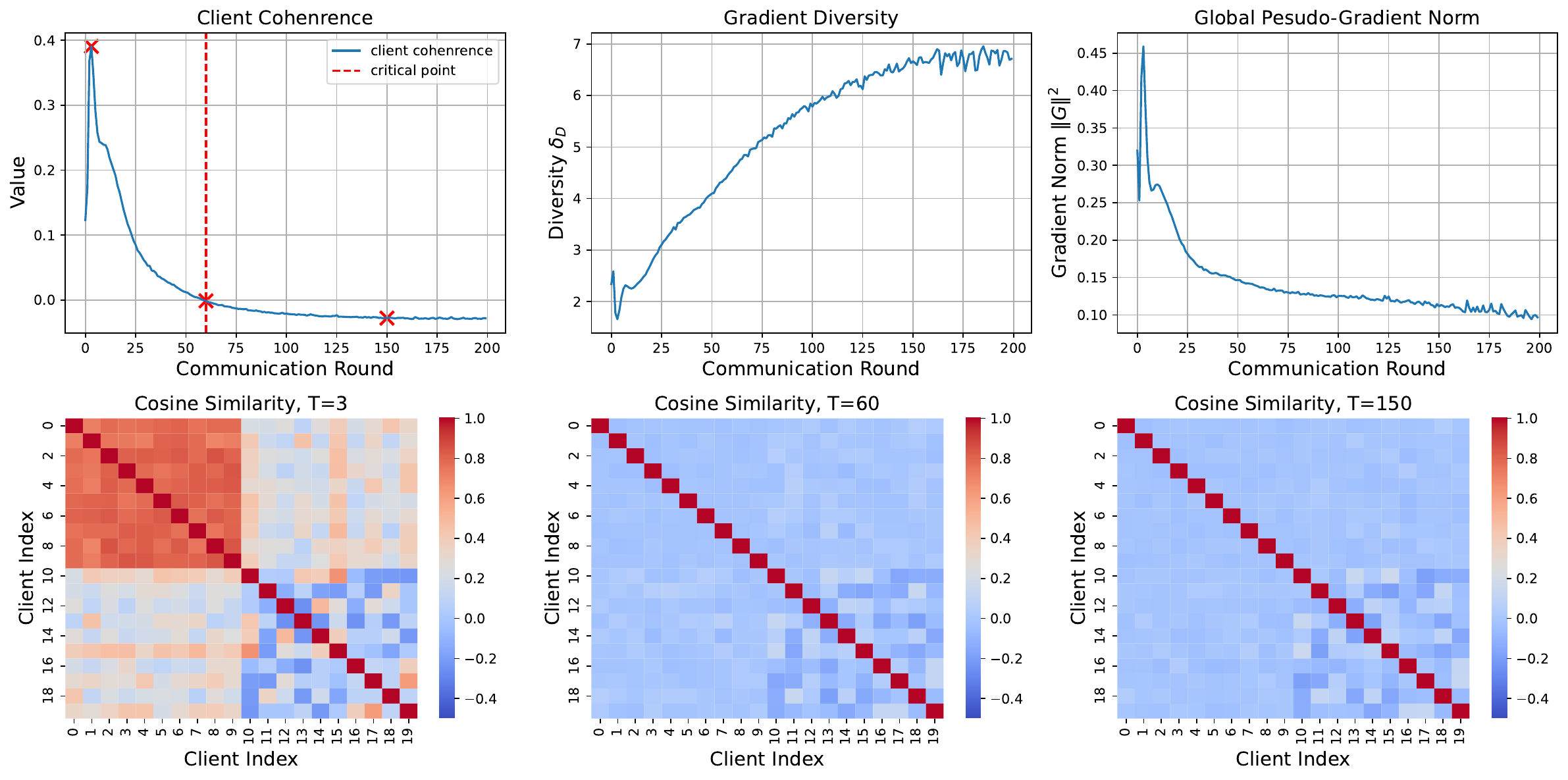}
\caption{Visualization of client coherence and cosine similarity.}\label{fig:fedlaw}
\end{figure}

\textbf{Relation between client coherence and LUD  } 
To illustrate the nuances, we roughly reproduced the experiments from Figure 5 in FedLAW~\citep{li2023revisiting}, as shown in Figure~\ref{fig:fedlaw}. The experiments involve 20 clients, where the first 10 have label-balanced datasets, and the remaining 10 have label-unbalanced datasets. We observe consistent trends across three different measures, suggesting their interrelation. Additionally, we visualize cosine similarity matrices at training stages $T={3, 60, 150}$, which reveal that local updates become less similar in later training rounds.

\begin{figure}[h]
\centering
\includegraphics[width=0.9\linewidth]{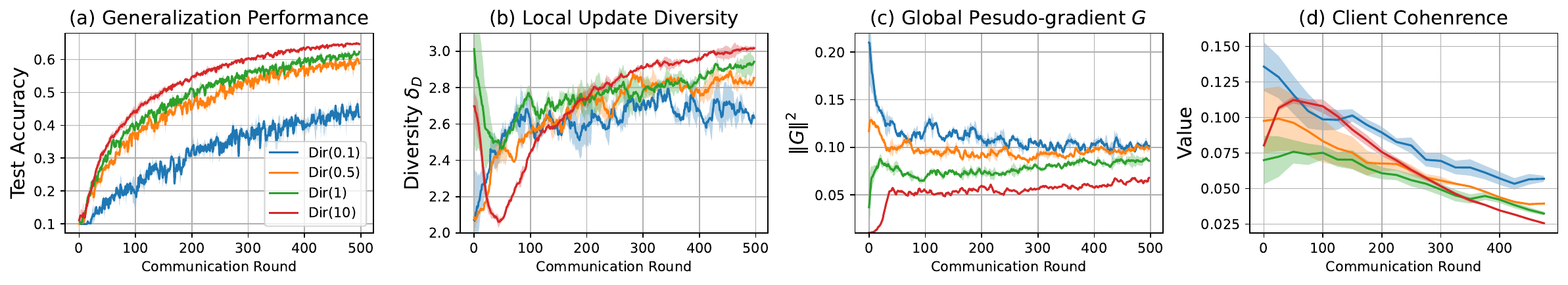}
\caption{Training dynamics of IID and Non-IID settings.}\label{fig:iid_comparison}
\end{figure}

\textbf{Discussion: the meaning of cosine similarity in FL.}
Cosine similarity plays a different role in FL training. We outline the varying meanings of consistent gradients (cosine values) during the training process:
\begin{itemize}
    \item  \textbf{Initialization and early stage:} High cosine values indicate similar data distributions (even IID) in the FL system~\citep{zeng2023stochastic, sattler2020clustered}. It is well-established that distributed learning with IID data outperforms non-IID data in terms of generalization. In this phase, client coherence suggests that \textit{gradient coherence claims that higher similarities between the gradients of samples will boost generalization}.
    
    \item \textbf{Training stage:} As training progresses, cosine similarity reflects conflicts between local objectives. Increased dissimilarity in gradients signals divergence from local minima. We observe that cosine values generally decrease over time, signaling growing gradient dissimilarity~\citep{sattler2020clustered, li2023revisiting}. Interestingly, we find that larger dissimilarity in later training rounds correlates with better generalization. This trend is further supported in Figure~\ref{fig:iid_comparison}, where we compare client coherence and LUD across different heterogeneous settings.
\end{itemize}
In summary, we clarify that the findings on client coherence~\citep{li2023revisiting, Chatterjee20} and gradient diversity~\citep{yin2018gradient} in federated learning are not contradictory.

\textbf{Observation experiments on IID settings } 
Figure~\ref{fig:iid_comparison} compares Dirichlet partitions Dir(1) and Dir(10) as IID settings with Dir(0.1) and Dir(0.5) as Non-IID settings. We observe distinct trends between the two. In IID settings, the global gradient norm remains lower than in Non-IID settings. However, the LUD curve is higher in IID settings, as the global norm stabilizes while local gradient norms increase due to stochasticity.
In essence, FL in IID settings behaves similarly to \textit{Local SGD}~\citep{DBLP:conf/iclr/GuLHA23}. Notably, the decaying property~\ref{asp:client_consunsus} of global norms may not hold in IID settings. Despite that, our convergence analysis of Theorem~\ref{theorem:fedavg_consensus} also holds, as the cumulative client consensus remains lower in IID settings.
Furthermore, the relationship between LUD and generalization continues to align with the theoretical findings on gradient diversity~\citep{yin2018gradient}.

\end{document}